\documentclass[10pt,twocolumn,letterpaper]{article}

\usepackage[pagenumbers]{cvpr} 
\usepackage{graphicx}
\usepackage{amsmath}
\usepackage{amssymb}
\usepackage{booktabs}
\usepackage{multirow}
\usepackage{paralist}

\usepackage[pagebackref,breaklinks,colorlinks]{hyperref}

\usepackage[capitalize]{cleveref}
\crefname{section}{Sec.}{Secs.}
\Crefname{section}{Section}{Sections}
\Crefname{table}{Table}{Tables}
\crefname{table}{Tab.}{Tabs.}

\begin{document}
\title{\LaTeX\ Author Guidelines for \confName~Proceedings}
\title{M$^{6}$Doc: A Large-Scale Multi-Format, Multi-Type, Multi-Layout, Multi-Language, Multi-Annotation Category Dataset for \\ Modern Document Layout Analysis}

\author{Hiuyi Cheng\textsuperscript{1}, Peirong Zhang\textsuperscript{1}, Sihang Wu\textsuperscript{2}, Jiaxin Zhang\textsuperscript{1}, \\Qiyuan Zhu\textsuperscript{2}, Zecheng Xie\textsuperscript{2}, Jing Li\textsuperscript{2}, Kai Ding\textsuperscript{3}, and Lianwen Jin\textsuperscript{1}\textsuperscript{*}\\
\textsuperscript{1}South China University of Technology\\
\textsuperscript{2}Huawei Cloud Computing Technologies Co., Ltd.\\ 
\textsuperscript{3}IntSig Information Co., Ltd.\\
\{eechenghiuyi, eeprzhang, msjxzhang\}@mail.scut.edu.cn, eelwjin@scut.edu.cn, danny\_ding@intsig.net, \\ 
\{wusihang2, zhuqiyuan2, xiezecheng1, lijing260\}@huawei.com}
\maketitle
\renewcommand{\thefootnote}{}
\footnotetext{\textsuperscript{*}Corresponding Author.}
\renewcommand{\thefootnote}{\arabic{footnote}}
\begin{abstract}
    Document layout analysis is a crucial prerequisite for document understanding, including document retrieval and conversion. Most public datasets currently contain only PDF documents and lack realistic documents. Models trained on these datasets may not generalize well to real-world scenarios. Therefore, this paper introduces a large and diverse document layout analysis dataset called $M^{6}Doc$. The $M^6$ designation represents six properties: (1) Multi-Format (including scanned, photographed, and PDF documents); (2) Multi-Type (such as scientific articles, textbooks, books, test papers, magazines, newspapers, and notes); (3) Multi-Layout (rectangular, Manhattan, non-Manhattan, and multi-column Manhattan); (4) Multi-Language (Chinese and English); (5) Multi-Annotation Category (74 types of annotation labels with 237,116 annotation instances in 9,080 manually annotated pages); and (6) Modern documents. Additionally, we propose a transformer-based document layout analysis method called TransDLANet, which leverages an adaptive element matching mechanism that enables query embedding to better match ground truth to improve recall, and constructs a segmentation branch for more precise document image instance segmentation. We conduct a comprehensive evaluation of $M^{6}Doc$ with various layout analysis methods and demonstrate its effectiveness. TransDLANet achieves state-of-the-art performance on $M^{6}Doc$ with 64.5\% mAP. The $M^{6}Doc$ dataset will be available at \url{https://github.com/HCIILAB/M6Doc}.
\end{abstract}
\begin{figure}[!htbp]
    \centering
    \includegraphics[width=\linewidth]{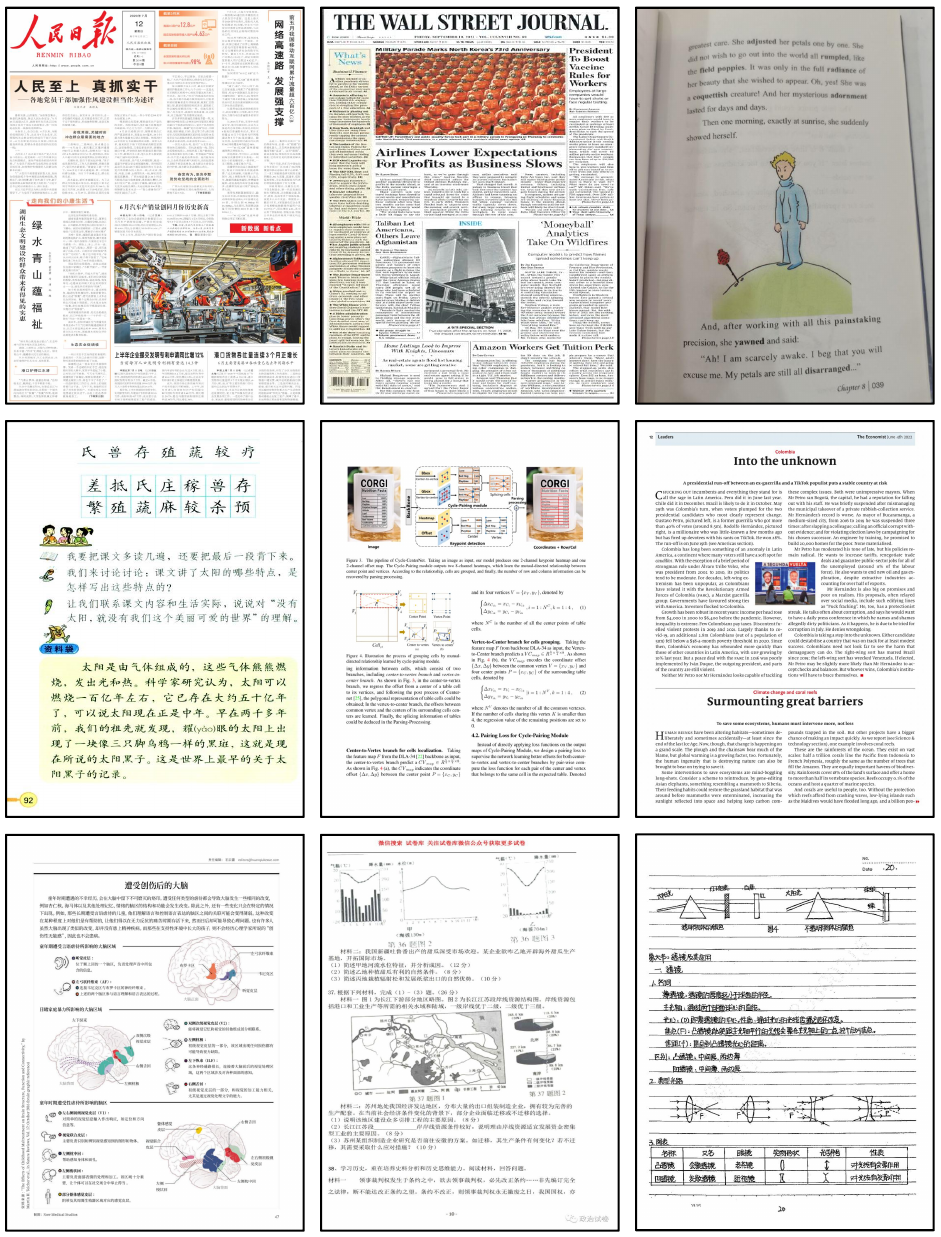}
    \caption{Examples of complex page layouts across different document formats, types, layouts, languages.}
    \label{fig1}
\end{figure}
\vspace{-1cm}
\section{Introduction}
\label{sec:intro}
    Document layout analysis (DLA) is a fundamental preprocessing task for modern document understanding and digitization, which has recently received increasing attention~\cite{article0}. DLA can be classified into physical layout analysis and logical layout analysis~\cite{article17}. Physical layout analysis considers the visual presentation of the document and distinguishes regions with different elements such as text, image, and table. Logical layout analysis distinguishes the semantic structures of documents according to the meaning and assigns them to different categories, such as chapter heading, section heading, paragraph, and figure note.

    Currently, deep learning methods have dominated DLA, which require a plethora of training data. Some datasets have been proposed in the community to promote the development of DLA, as shown in Table~\ref{tab1}. However, these datasets have several limitations. (1) Small size. Early DLA datasets, such as PRImA~\cite{article1} and DSSE200~\cite{article5}, were small-scale and contained only hundreds of images. (2) Limited document format. The formats of current public large-scale datasets such as PubLayNet~\cite{article2}, DocBank~\cite{article3}, and DocLayNet~\cite{article4}, are all PDF documents. It presents a huge challenge to evaluate the effectiveness of different methods in realistic scenarios. (3) Limited document diversity. Most datasets include only scientific articles, which are typeset using uniform templates and severely lack variability. Although DocLayNet~\cite{article5} considers documents of seven types, they are not commonly used. The lack of style diversity would prejudice the development of multi-domain general layout analysis. (4) Limited document languages. Most datasets' language is English. Since the text features of documents in different languages are fundamentally different, DLA methods may encounter domain shift problems in different languages, which remain unexplored. (5) Few annotation categories. The annotation categories of current datasets are not sufficiently fine-grained, preventing more granular layout information extraction.

    To promote the development of fine-grained logical DLA in realistic scenarios, we have built the Multi-Format, Multi-Type, Multi-Layout, Multi-Language, and Multi-Annotation Categories Modern document ($M^6Doc$) dataset. $M^6Doc$ possesses several advantages. Firstly, $M^6Doc$ considers three document formats (scanned, photographed, and PDF) and seven representative document types (scientific articles, magazines, newspapers, etc.). Since scanned/photographed documents are commonly seen and widely used, the proposed $M^6Doc$ dataset presents great diversity and closely mirrors real-world scenarios. Secondly, $M^6Doc$ contains 74 document annotation categories, which are the most abundant and fine-grained up to date. Thirdly, $M^6Doc$ is the most detailed manually annotated DLA dataset, as it contains 237,116 annotation instances in 9,080 pages. Finally, $M^6Doc$ includes four layouts (rectangular, Manhattan, non-Manhattan, and multi-column Manhattan) and two languages (Chinese and English), covering more comprehensive layout scenarios. Several examples of the $M^6Doc$ dataset are shown in Figure~\ref{fig1}.

    In addition, we propose a transformer-based model, TransDLANet, to perform layout extraction in an instance segmentation manner effectively. It adopts a standard Transformer encoder without positional encoding as a feature fusion method and uses an adaptive element matching mechanism to enable the query vector to better focus on the unique features of layout elements. This helps understand the spatial and global interdependencies of distinct layout elements and also reduces duplicate attention on the same instance. Subsequently, a dynamic decoder is exploited to perform the fusion of RoI features and image features. Finally, it uses three parameter-shared multi-layer perception (MLP) branches to decode the fused interaction features for multi-task learning.

    The contributions of this paper are summarized as follows:
    
    \begin{itemize}
        \item[$\bullet$] $M^{6}Doc$ is the first layout analysis dataset that contains both real-world (photographed and scanned) files and born-digital files. Additionally, it is the first dataset that includes Chinese examples. It has several representative document types and layouts, facilitating the development of generic layout analysis methods.
        \item[$\bullet$] $M^{6}Doc$ is the most fine-grained logical layout analysis categories. It can serve as a benchmark for several related tasks, such as logical layout analysis, formula recognition, and table analysis.
        \item[$\bullet$] We propose the TransDLANet, a Transformer-based method for document layout analysis. It includes a Transformer-like encoder to better capture the correlation between queries, a dynamic interaction decoder, and three multi-ayer perceptron branches with shared parameters to decode the fused interaction features for multi-task learning.
    \end{itemize}
    
    \begin{table*}[!htbp]
        \vspace{-1cm}
        \renewcommand{\arraystretch}{1}
        \caption{Modern Document Layout Analysis Datasets. \textbf{A.M.} denotes the annotating means.}
        \label{tab1}
        \resizebox{\textwidth}{!}{
            \begin{tabular}{cccccccc}
            \toprule
            \textbf{Dataset} & \textbf{\#Image} & \textbf{\#Class} & \textbf{\#Instance} & \textbf{A.M.} & \textbf{Format} & \textbf{Document Type} & \textbf{Language} \\ \hline
            DSSE200~\cite{article5} & 200 & 6 & - & Automatic & PDF & Magazines, Academic papers. & English \\ \hline
            DAD~\cite{article6} & 5,980 & 5 & 90,923 & Automatic & PDF & Articles & English \\ \hline
            PubMed~\cite{article9} & 12,871 & 5 & 257,830 & Automatic & PDF & Articles & English \\ \hline
            Chn~\cite{article9} & 8,005 & 5 & 203,456 & Automatic & PDF & Chinese Wikipedia pages & Chinese \\ \hline
            PubLayNet~\cite{article2} & 360K & 5 & 3,311,660 & Automatic & PDF & Articles & English \\ \hline
            DocBank~\cite{article3} & 500K & 13 & - & Automatic & PDF & Articles & English \\ \hline
            DocLayNet~\cite{article4} & 80,863 & 11 & 1,107,470 & Manual & PDF & \begin{tabular}[c]{@{}c@{}}Financial Reports, Manuals, \\ Scientific Articles, Laws \& Regulations, \\ Patents, Government Tenders.\end{tabular} & \begin{tabular}[c]{@{}c@{}}English, German, \\ French, Japanese\end{tabular} \\ \hline
            PRImA~\cite{article1} & 305 & 10 & - & Automatic & Scanned & \begin{tabular}[c]{@{}c@{}}Magazine, Technical article, Forms, \\ Bank statements, Advertisements\end{tabular} & English \\ \hline
            BCE-Arabic-v1~\cite{article7} & 1,833 & 3 & - & Automatic & Scanned & Arabic books & Arabic \\ \hline
            BCE-Arabic-v2~\cite{article8} & 9,000 & 21 & - & Automatic & Scanned & Arabic books & Arabic \\ \hline
            M$^6$Doc (\textbf{Ours}) & 9,080 & 74 & 237,116 & Manual & \begin{tabular}[c]{@{}c@{}}PDF, Scanned, \\ Photographed \end{tabular} & \begin{tabular}[c]{@{}c@{}} Scientific articles, Textbooks, \\ Books, Test papers, Magazines,\\ Newspapers, Notes\end{tabular} & English, Chinese \\
            \bottomrule
        \end{tabular}}
    \end{table*}
\section{Related Works}
\subsection{Modern Layout Analysis Dataset}
    A variety of modern layout analysis datasets have been created in recent years. In 2009, Antonacopoulos et al.~\cite{article1} presented the PRImA dataset, which was the first commonly used real-world dataset with 305 images of magazines and scientific articles. In 2019, Zhong et al.~\cite{article2} published the PubLayNet dataset, which contains over 360,000 page samples annotated with typical document layout elements such as text, heading, list, graphic, and table. Annotations were automatically generated by matching PDFs and XML formats of articles from the PubMed Central Open Access subset. In 2020, researchers at Microsoft Research Asia built the DocBank dataset~\cite{article3}, which contains 500,000 document pages and fine-grained token-level annotations for document layout analysis. It was developed based on a large number of PDF files of papers compiled by the LaTeX tool. Unlike the conventional manual annotating process, they approach obtaining high-quality annotations using a weakly supervised approach in a simple and efficient manner. In 2022, IBM researchers presented the DocLayNet dataset~\cite{article4}, which contains 80,863 manually annotated pages. It contains six document types (technical manuals, annual company reports, legal text, and government tenders), 11 categories of annotations, and four languages (English documents close to 95\%). A few pages in the DocLayNet dataset have multiple manual annotations, which allows for experiments in annotation uncertainty and quality control analysis.

    However, the predominant document format for large datasets is PDF, not scanned and photographed images as in real-world scenarios. Only a few public datasets include real-world data. The variety of layouts in current public datasets is still very limited and is not conducive to the development of logical layout analysis. Currently, 95\% of the publicly available datasets are English documents, which are largely unsuitable for Asian language documents. To this end, we propose the $M^6Doc$ dataset to facilitate the development of layout analysis.

    \subsection{Deep Learning for Layout Analysis}
    Earlier layout analysis methods~\cite{article10, article11, article13, article14, article15} used rule-based and heuristic algorithms, so they were limited to applications on certain simple types of documents, and the generalization performance of such methods was poor. However, with the development of deep learning, DLA methods based on deep learning have been developed to tackle challenging tasks. Mainstream approaches include object detection-based models~\cite{article9, article49, article50}, segmentation-based models\cite{article17, article51, article52}, and multi-modal methods\cite{article18, article53, article5}. For example, Li et al.~\cite{article9} considered DLA as an object detection task and added a domain adaptation module to study cross-domain document object detection tasks. Lee et al.~\cite{article17} used segmentation methods to solve DLA problems and introduced trainable multiplication layer techniques for improving the accuracy of object boundary detection to improve the performance of pixel-level segmentation networks. Zhang et al.~\cite{article18} proposed a unified framework for multi-modal layout analysis by introducing semantic information in a new semantic branch of Mask R-CNN\cite{article35} and a module for modeling element relationships. Behind their success, large datasets are required for training and evaluating the models.

    However, the lack of a multi-format, multi-type, multi-language, and multi-label categorized logical layout analysis dataset makes it difficult for current methods to obtain good results in real-world and other language scenarios. Moreover, a data format that links visual and textual features has not yet been established for multi-modal tasks.
    \begin{table*}[!htbp]
        \vspace{-1cm}
        \caption{$M^6Doc$ dataset overview.}
        \label{tab2}
        \resizebox{\textwidth}{!}{
            \begin{tabular}{lcccccc|lcccccc}
                \hline
                \multicolumn{1}{l}{\multirow{2}{*}{\textbf{Category}}} & \multicolumn{2}{c}{\textbf{Training}} & \multicolumn{2}{c}{\textbf{Validate}} & \multicolumn{2}{c|}{\textbf{Test}} & \multicolumn{1}{l}{\multirow{2}{*}{\textbf{Category}}} & \multicolumn{2}{c}{\textbf{Training}} & \multicolumn{2}{c}{\textbf{Validate}} & \multicolumn{2}{c}{\textbf{Test}} \\
                \multicolumn{1}{c}{} & \textbf{Number} & \textbf{\%} & \textbf{Number} & \textbf{\%} & \textbf{Number} & \textbf{\%} & \multicolumn{1}{c}{} & \textbf{Number} & \textbf{\%} & \textbf{Number} & \textbf{\%} & \textbf{Number} & \textbf{\%} \\ \hline
                \_background\_ & 0 & 0.000 & 0 & 0.000 & 0 & 0.000 & institute & 60 & 0.042 & 9 & 0.039 & 28 & 0.040 \\
                QR code & 59 & 0.041 & 15 & 0.065 & 23 & 0.032 & jump line & 381 & 0.266 & 63 & 0.271 & 180 & 0.254 \\
                advertisement & 257 & 0.180 & 45 & 0.194 & 145 & 0.205 & kicker & 516 & 0.361 & 91 & 0.392 & 257 & 0.363 \\
                algorithm & 12 & 0.008 & 3 & 0.013 & 12 & 0.017 & lead & 664 & 0.464 & 109 & 0.470 & 285 & 0.402 \\
                answer & 165 & 0.115 & 30 & 0.129 & 77 & 0.109 & marginal note & 238 & 0.166 & 37 & 0.159 & 101 & 0.143 \\
                author & 2,424 & 1.695 & 403 & 1.736 & 1,188 & 1.676 & matching & 7 & 0.005 & 1 & 0.004 & 8 & 0.011 \\
                barcode & 10 & 0.007 & 1 & 0.004 & 3 & 0.004 & mugshot & 73 & 0.051 & 11 & 0.047 & 46 & 0.065 \\
                bill & 3 & 0.002 & 2 & 0.009 & 3 & 0.004 & option & 3,198 & 2.236 & 515 & 2.219 & 1,577 & 2.225 \\
                blank & 189 & 0.132 & 58 & 0.250 & 90 & 0.127 & ordered list & 1,012 & 0.707 & 172 & 0.741 & 510 & 0.720 \\
                bracket & 863 & 0.603 & 164 & 0.707 & 273 & 0.385 & other question number & 42 & 0.029 & 3 & 0.013 & 31 & 0.044 \\
                breakout & 411 & 0.287 & 72 & 0.310 & 188 & 0.265 & page number & 4,782 & 3.343 & 803 & 3.460 & 2,383 & 3.363 \\
                byline & 1,276 & 0.892 & 185 & 0.797 & 660 & 0.931 & paragraph & 65,642 & 45.891 & 10,575 & 45.562 & 33,069 & 46.664 \\
                caption & 3,508 & 2.452 & 605 & 2.607 & 1,766 & 2.492 & part & 524 & 0.366 & 89 & 0.383 & 283 & 0.399 \\
                catalogue & 39 & 0.027 & 10 & 0.043 & 19 & 0.027 & play & 10 & 0.007 & 3 & 0.013 & 2 & 0.003 \\
                chapter title & 245 & 0.171 & 33 & 0.142 & 124 & 0.175 & poem & 98 & 0.069 & 18 & 0.078 & 33 & 0.047 \\
                code & 62 & 0.043 & 7 & 0.030 & 31 & 0.044 & reference & 149 & 0.104 & 23 & 0.099 & 62 & 0.087 \\
                correction & 9 & 0.006 & 1 & 0.004 & 6 & 0.008 & sealing line & 3 & 0.002 & 2 & 0.009 & 5 & 0.007 \\
                credit & 1,523 & 1.065 & 255 & 1.099 & 728 & 1.027 & second-level question number & 2,773 & 1.939 & 377 & 1.624 & 1,330 & 1.877 \\
                dateline & 901 & 0.630 & 140 & 0.603 & 482 & 0.680 & second-level title & 273 & 0.191 & 48 & 0.207 & 140 & 0.198 \\
                drop cap & 414 & 0.289 & 71 & 0.306 & 234 & 0.330 & section & 2,508 & 1.753 & 408 & 1.758 & 1,228 & 1.733 \\
                editor's note & 39 & 0.027 & 4 & 0.017 & 9 & 0.013 & section title & 897 & 0.627 & 171 & 0.737 & 442 & 0.624 \\
                endnote & 35 & 0.024 & 4 & 0.017 & 19 & 0.027 & sidebar & 54 & 0.038 & 10 & 0.043 & 27 & 0.038 \\
                examinee information & 8 & 0.006 & 2 & 0.009 & 6 & 0.008 & sub section title & 567 & 0.396 & 107 & 0.461 & 269 & 0.380 \\
                fifth-level title & 13 & 0.009 & 2 & 0.009 & 20 & 0.028 & subhead & 1,998 & 1.397 & 394 & 1.698 & 1,069 & 1.508 \\
                figure & 7,614 & 5.323 & 1,242 & 5.351 & 3,762 & 5.309 & subsub section title & 101 & 0.071 & 21 & 0.090 & 71 & 0.100 \\
                first-level question number & 5,669 & 3.963 & 930 & 4.007 & 2,740 & 3.866 & supplementary note & 986 & 0.689 & 158 & 0.681 & 487 & 0.687 \\
                first-level title & 586 & 0.410 & 81 & 0.349 & 292 & 0.412 & table & 821 & 0.574 & 146 & 0.629 & 409 & 0.577 \\
                flag & 30 & 0.021 & 5 & 0.022 & 12 & 0.017 & table caption & 287 & 0.201 & 41 & 0.177 & 143 & 0.202 \\
                folio & 1,442 & 1.008 & 213 & 0.918 & 685 & 0.967 & table note & 8 & 0.006 & 2 & 0.009 & 5 & 0.007 \\
                footer & 1,984 & 1.387 & 310 & 1.336 & 987 & 1.393 & teasers & 32 & 0.022 & 7 & 0.030 & 7 & 0.010 \\
                footnote & 295 & 0.206 & 49 & 0.211 & 139 & 0.196 & third-level question number & 240 & 0.168 & 36 & 0.155 & 102 & 0.144 \\
                formula & 1,3090 & 9.151 & 2,058 & 8.867 & 6,191 & 8.736 & third-level title & 146 & 0.102 & 44 & 0.190 & 94 & 0.133 \\
                fourth-level section title & 15 & 0.010 & 3 & 0.013 & 19 & 0.027 & title & 201 & 0.141 & 35 & 0.151 & 100 & 0.141 \\
                fourth-level title & 70 & 0.049 & 13 & 0.056 & 66 & 0.093 & translator & 73 & 0.051 & 11 & 0.047 & 38 & 0.054 \\
                header & 1,877 & 1.312 & 297 & 1.280 & 969 & 1.367 & underscore & 3,687 & 2.578 & 590 & 2.542 & 1,717 & 2.423 \\
                headline & 4,115 & 2.877 & 643 & 2.770 & 1,981 & 2.795 & unordered list & 497 & 0.347 & 84 & 0.362 & 271 & 0.382 \\
                index & 214 & 0.150 & 36 & 0.155 & 100 & 0.141 & weather forecast & 10 & 0.007 & 3 & 0.013 & 3 & 0.004 \\
                inside & 16 & 0.011 & 1 & 0.004 & 5 & 0.007 & \textbf{Total} & \textbf{143,040} & \textbf{100} & \textbf{23,210} & \textbf{100} & \textbf{70,866} & \textbf{100} \\ \hline
            \end{tabular}}
    \end{table*}
\vspace{-0.5cm}
\section{$M^6Doc$ Dataset}
    \footnotetext[1]{\url{https://arxiv.org/}}
    \footnotetext[2]{\url{http://paper.people.com.cn/}}
    \footnotetext[3]{\url{https://vk.com/}}
    The $M^6Doc$ dataset contains a total of 9,080 modern document images, which are categorized into seven subsets, \emph{i.e.}, scientific article (11\%), textbook (23\%), test paper (22\%), magazine (22\%), newspaper (11\%), note (5.5\%), and book (5.5\%) according to their content and layouts. It contains three formats: PDF (64\%), photographed documents (5\%), and scanned documents (31\%). The dataset includes a total of 237,116 annotated instances.

    The $M^{6}Doc$ datasets were collected from various sources, including arXiv\footnotemark[1], the official website of the Chinese People's Daily\footnotemark[2], and VKontakte\footnotemark[3]. The source and composition of different subsets are shown below.

    \begin{itemize}
        \item[$\bullet$] The scientific article subset includes articles obtained by searching with the keywords ``Optical Character Recognition'' and ``Document Layout Analysis'' on arXiv. PDF files were then downloaded and converted to images.
        \item[$\bullet$] The textbook subset contains 2,080 scanned document images from textbooks for three grades (elementary, middle, and high school) and nine subjects (Chinese, Math, English, Physics, Chemistry, Biology, History, Geography, and Politics).
        \item[$\bullet$] The test paper subset consists of 2,000 examination papers covering the same nine subjects as the textbook subset.
        \item[$\bullet$] The magazine subset includes 1,000 Chinese and English magazines in PDF format, respectively. The Chinese magazines were sourced from five publishers: Global Science, The Mystery, Youth Digest, China National Geographic, and The Reader. The English magazines were sourced from five American publishers: The New Yorker, New Scientist, Scientific American, The Economist, and Time USA.
        \item[$\bullet$] The newspaper subset contains 500 PDF document images from the Chinese People's Daily and the Wall Street Journal.
        \item[$\bullet$] The note subset consists of students' handwritten notes in nine subjects, including 500 scanned pages.
        \item[$\bullet$] The book subset contains 500 photographed images, which were acquired from 50 books with 10 pages each. Each book has a distinct layout, resulting in considerable diversity in this subset.
    \end{itemize}

    For a fair evaluation, we divided the dataset into training, validation, and test sets in a ratio of 6:1:3. We also ensured that the different labels were in equal proportions in the three sets. Table \ref{tab2} summarizes the overall frequency and distribution of labels in different sets.
    \begin{figure*}[!tbp]
        \vspace{-1cm}
        \centering
        \includegraphics[width=0.8\textwidth, height=0.25\textwidth]{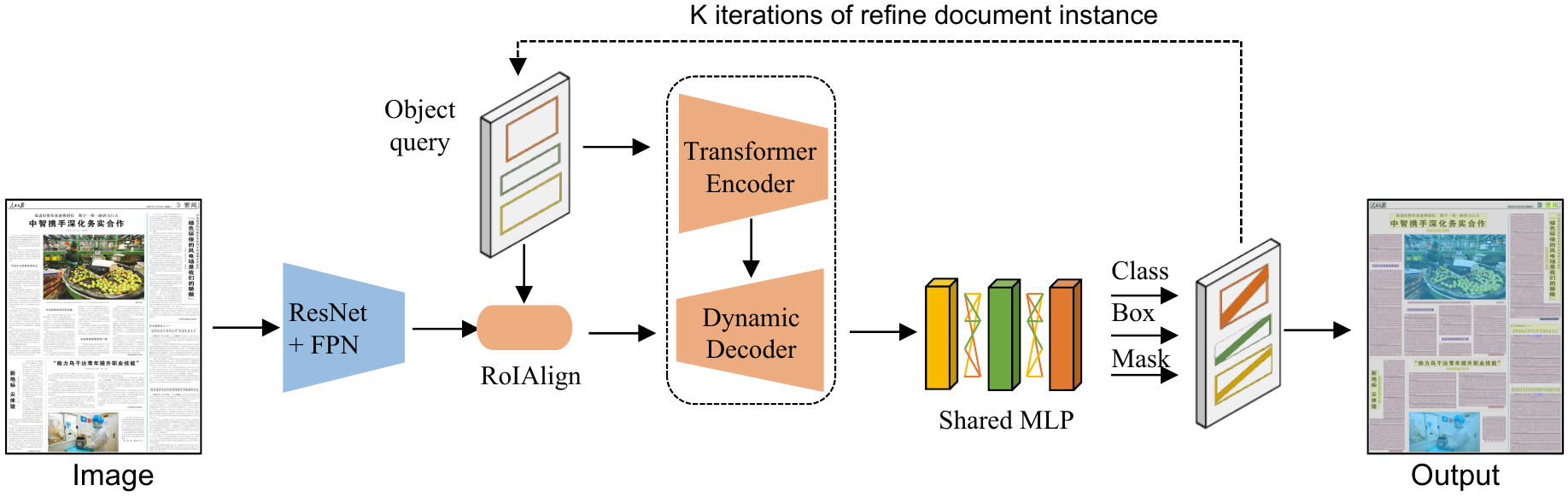}
        \caption{The pipeline of TransDLANet contains four main components: 1) a CNN-based backbone; 2) a transformer encoder; 3) a dynamic decoder that decodes the instance-level features; and 4) three shared multi-layer perceptron(MLP) branches that obtain the classification confidence, bounding boxes, and segmentation mask of the document instance region.}
        \label{fig2}
    \end{figure*}
\section{Data Annotation}
    \footnotetext[1]{\url{https://www.youtube.com/watch?v=7sSJtScnsjE}}
    \footnotetext[2]{\url{https://www.youtube.com/watch?v=LcsOuGcaqZs}}
    \textbf{Label definition}. To ensure that the definition of document layout elements is reasonable and traceable, we reviewed relevant information, such as layout knowledge and layout design. We also used knowledge from the book "Page Design: New Layout \& Editorial Design(2019)"\cite{book1} and referred to YouTube video explanations regarding magazine\footnotemark[1] and newspaper\footnotemark[2] layouts. In most cases, we followed the Wikipedia definition. Consequently, we defined 74 detailed document annotation labels. The key factors in selecting these annotation labels include (1) the commonality of annotation labels between different document types, (2) the specificity of labels between different document types, (3) the frequency of labels, and (4) the recognition of independent pages. We first unified the labels between different documents to the maximum extent and then defined the labels for certain document types for differentials. Commonality and specificity ensure that the defined labels can adapt to multiple document types, which implies that a more detailed logical layout analysis for a certain type of document can be performed. It differs from how labels are defined in DocBank, PubLayNet, and DocLayNet, which all ignore defining specific labels for different document types.

    \textbf{Annotation guideline}. We provide a detailed annotation guideline (over 170 pages) and some typical annotation examples. 47 annotators performed the annotation task strictly according to the guidelines.

    Several key points of the guideline that are different from DocBank, PubLayNet, and DocLayNet are summarized as follows:
    \begin{itemize}
        \item[$\bullet$] We distinguish table caption and figure caption into two categories.
        \item[$\bullet$] We distinguish the ordered list and unordered list into two categories.
        \item[$\bullet$] All list-items are grouped together into one list object. This definition differs from DocLayNet, which considers single-line elements as list-item if the list-item are paragraphs with hanging indentation.
        \item[$\bullet$] Bold emphasized text at the beginning of a paragraph is not considered a heading unless it appears on a separate line or with heading formatting, such as 1.1.1.
        \item[$\bullet$] The headings at different levels are defined in detail.
        \item[$\bullet$] The formulas inside the paragraphs are marked.
    \end{itemize}

    The annotation results showed that different annotators interpreted ambiguous scenarios differently, such as (1) in the absence of obvious borders, it is sometimes difficult to determine whether a region is a table or a paragraph; (2) whether images with sub-images should be annotated separately or holistically; and (3) in the absence of obvious markers or separators, it is sometimes difficult to determine whether a paragraph is a list item or a body. It was difficult to unify the consistency of the results of the 47 annotators. Therefore, we provided consistent annotation requirements for ambiguous scenarios. To further ensure the consistency of the annotation results, all data were finally checked by the author. The annotation files followed the MS COCO annotation format\cite{article19} for object detection. Detailed annotation guidelines and the $M^6Doc$ dataset will be available for reference. A more detailed labeling process is provided in the Supplementary Material.
    
    \vspace{-0.2cm}
    \section{TransDLANet}
    Our method closely follows the framework of ISTR~\cite{article20}, but differs at its core by leveraging an adaptive element matching mechanism that enables query embedding to better match ground truth and improve recall. We use the transformer encoder as a characteristic fusion method without position encoding and construct a segmentation branch for more precise document image instance segmentation. Additionally, we use three multi-layer perception(MLP) branches with shared parameter for multi-task learning.

    \textbf{TransDLANet architecture}. The overall architecture is depicted in Figure \ref{fig2}. We use a CNN-based backbone to extract document image features, and RoIAlign to extract the image features for the pre-defined query vectors. The Transformer encoder performs self-attentive feature learning on query embedding vectors and uses an adaptive element matching mechanism to enhance further the association between document instances encoded by the query vectors. The dynamic interaction-based decoding module (Dynamic Decoder) fuses the query vector with the features of the bounding box image region obtained by the query vector through the RoIAlign. Three shared parameter MLP branches are used for decoding the classification confidence, the bounding boxes' coordinate position, and the segmentation mask of the document instance region. Finally, we repeat this process for K iterations to refine the final document instance.

    \vspace{-0.2cm}
    \section{Experiment}
    \vspace{-0.1cm}
    \subsection{Datasets}
    Our experiments are conducted on a number of commonly known document layout analysis benchmarks, including {DocBank}~\cite{article3}, PubLayNet~\cite{article2}, and DocLayNet~\cite{article4}.
     
     
    
    \subsection{Implementation Details}
    We adopted ResNet-101 pretrained on ImageNet\cite{article43} as our model's backbone. We used the AdamW optimizer\cite{article44} to train the model, setting the base learning rate to $2 \times 10^{-5}$. The default training epoch was set to 500, and the learning rates descended to $2 \times 10^{-6}$ and $2 \times 10^{-7}$ at 50\% and 75\% of the training epochs, respectively. During training, we used random crop augmentations and scaled the input images such that the shortest side was at least 704-896 pixels and the longest side was at most 1333 pixels to ensure optimal performance.
    \subsection{Significance of $M^{6}Doc$}
    Due to the inconsistency in labeling across different datasets, it is not feasible to directly compare the mAP scores. Consequently, we have used visualization results to perform our analysis. The Supplementary Material includes the results of qualitative experiments in which we mapped the labels of $M^6Doc$ to labels of other datasets.
    \begin{figure*}[!tbp]
        \vspace{-1cm}
        \centering
        \includegraphics[width=\textwidth]{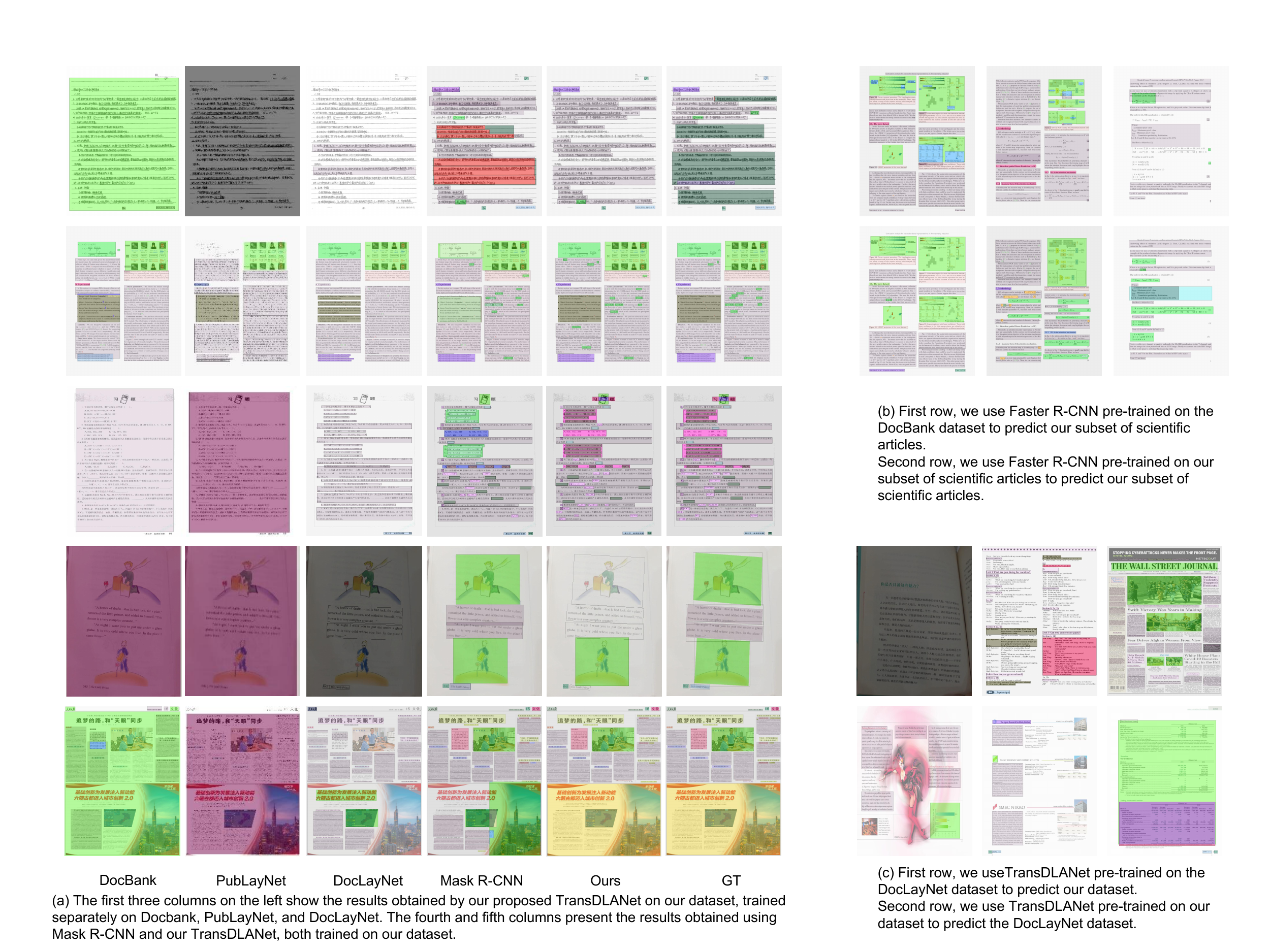}
        \caption{Visualization results. Zoom in for better view.}
        \label{fig3}
    \end{figure*}
    
    \textbf{Significance of format diversity}. Models trained on existing benchmark datasets such as DocBank, DocLayNet, and PubLayNet are not effective in processing some novel scenarios proposed in $M^{6}Doc$, such as scanned and photographed images. The specific analysis is as follows: the first row of the first three columns in Figure~\ref{fig3} (a) demonstrates that models trained on DocBank, DocLayNet, and PubLayNet are not effective in identifying document instances in scanned handwritten notes, likely due to the differences between handwritten and printed documents. Rows 3 and 4 of the first two columns reveal that the models trained on DocBank and PubLayNet are unable to process the scanned textbook and photographed book datasets, likely due to the complex backgrounds and tilting and brightness variation phenomena in these images. However, models trained on $M^{6}Doc$ well handle scanned and photographed images, as shown in columns 4-6 of Figure~\ref{fig3} (a). These results suggest that providing a training set containing scanned and photographed images is crucial for developing models that can handle diverse document formats.
  
    \textbf{Significance of type diversity}. The importance of type diversity is demonstrated in Figure~\ref{fig3} (a), where models trained on DocBank and PubLayNet fail to understand layouts for the new document types (note, textbook, book, newspaper) introduced in $M^{6}Doc$. This is due to the fact that DocBank and PubLayNet are limited to only one document type with restricted layouts. In contrast, $M^{6}Doc$ provides a diverse set of document types and complex layouts, which enables trained models to generalize well on DocBank and PubLayNet. Additionally, as seen in Figure~\ref{fig3} (c), DocLayNet and $M^{6}Doc$ have different data sources, resulting in significantly different layouts. As a result, models trained on $M^{6}Doc$ or DocLayNet do not perform well on each other. Hence, the need for diverse document types in a dataset is crucial for addressing generic layout analysis.

    \textbf{Significance of detailed labels}. The importance of detailed labels is demonstrated by comparing the performance of models trained on DocBank and the scientific article subset of $M^{6}Doc$, which have the same layout distribution. In our experiments, we used the Faster-RCNN model to predict the test set of the scientific article subset. Figure~\ref{fig3} (b) shows that the model trained on DocBank tends to detect large paragraphs of text while ignoring formulas, likely due to the large region of paragraph annotation used in DocBank. However, our model trained on the scientific article subset is able to avoid this issue and achieve more accurate segmentation results on the same test set by using more detailed labels. It's worth noting that the  scientific article subset only contains 600 images, yet adding more detailed labels improved the model's performance. This suggests that having more labels with fewer data may be more beneficial than having fewer labels with more data.

    \subsection{Comparisons with object detection and instance segmentation methods}
    In this section, we present the results of a thorough evaluation of $M^{6}Doc$ using different layout analysis techniques, which could serve as a benchmark for performance comparison. Further experiments on the performance of TransDLANet on nine sub-datasets of $M^{6}Doc$ are included in the Supplementary Material.
    
     We used RetianNet\cite{article28}, YOLOv3\cite{article29}, GFL\cite{article30}, FCOS\cite{article31}, FoveaBox\cite{article32}, Faster R-CNN\cite{article33}, Cascade R-CNN\cite{article34}, Mask R-CNN\cite{article35}, Cascade Mask R-CNN\cite{article34}, Deformable DETR\cite{article36}, and ISTR\cite{article37} as object detection baselines, while used HTC\cite{article38}, SCNet\cite{article39}, QueryInst\cite{article40}, SOLO\cite{article41}, and SOLOv2\cite{article42} as instance segmentation baselines to evaluate the $M^{6}Doc$ dataset. As the $M^{6}Doc$ dataset consists of different document types and layouts, and the instances have varying scales, it is challenging for anchor-based regression detection models to set up anchors that can fit all document instances. Therefore, the anchor ratios were adjusted to [0.0625, 0.125, 0.25, 0.5, 1.0, 2.0, 4.0, 8.0, 16.0] instead of the original three anchor ratios [0.5, 1.0, 2.0] for anchor-based models. For pure bounding box methods, the segmentation metrics were calculated using the detected bounding box as the segmentation mask. For pure instance segmentation methods, the minimum bounding rectangle was used to calculate the metrics for the bounding box. The results of the experiments are presented in the following sections.
    
    \begin{table}[!ht]
    \caption{Performance comparisons on $M^{6}Doc$.}
    \label{tab3}
        \resizebox{\linewidth}{!}{
        \begin{tabular}{l|c|cccc|ccc}
        \hline
            \multicolumn{1}{c|}{} & \multicolumn{1}{c|}{} & \multicolumn{4}{c|}{Object} & \multicolumn{3}{c}{Instance} \\
            \multicolumn{1}{c|}{} & \multicolumn{1}{c|}{} & \multicolumn{4}{c|}{Detection} & \multicolumn{3}{c}{Segmentation} \\ \cline{3-9}
            \multicolumn{1}{l|}{\multirow{-3}{*}{Method}} & \multicolumn{1}{l|}{\multirow{-3}{*}{Backbone}} & mAP & AP50 & AP75 & Recall & mAP & AP50 & AP75 \\ \hline
            RetinaNet\cite{article28} & ResNet-101 & 21.4 & 33.1 & 23.3 & 37.4 & 21.0 & 33.0 & 22.6 \\
            YOLOv3\cite{article29} & DarkNet-53 & 59.8 & 75.6 & 68.1 & 72.4 & - & - & - \\
            GFL\cite{article30} & ResNet-101 & 34.7 & 50.8 & 38.7 & 48.7 & 33.8 & 50.6 & 37.0 \\
            FCOS\cite{article31} & ResNet-101 & 40.6 & 59.3 & 45.9 & 59.5 & 39.3 & 58.9 & 43.1 \\
            FoveaBox\cite{article32} & ResNet-101 & 45.1 & 66.1 & 51.7 & 59.4 & 43.7 & 65.8 & 49.2 \\
            Faster R-CNN\cite{article33} & ResNet-101 & 49.0 & 67.8 & 57.2 & 57.2 & 47.8 & 67.8 & 55.2 \\
            Cascade R-CNN\cite{article34} & ResNet-101 & 54.1 & 70.4 & 62.3 & 61.4 & 52.7 & 70.2 & 60.1 \\
            Mask R-CNN\cite{article35} & ResNet-101 & 40.1 & 58.4 & 46.2 & 50.8 & 39.7 & 58.4 & 45.6 \\
            Cascade Mask R-CNN\cite{article34} & ResNet-101 & 54.4 & 70.5 & 62.9 & 62.1 & 52.9 & 70.4 & 60.6 \\
            HTC\cite{article38} & ResNet-101 & 58.2 & 74.3 & 67.2 & 68.1 & 57.1 & 74.4 & 65.7 \\
            SCNet\cite{article39} & ResNet-101 & 56.1 & 73.5 & 65.1 & 67.3 & 55.3 & 73.3 & 63.6 \\
            SOLO\cite{article41} & ResNet-101 & 38.7 & 56.0 & 42.7 & 54.9 & 38.7 & 56.3 & 43.0 \\
            SOLOv2\cite{article42} & ResNet-101 & 46.8 & 67.5 & 51.4 & 61.5 & 48.3 & 67.5 & 53.4 \\
            Deformable DETR\cite{article36} & ResNet-101 & 57.2 & 76.8 & 63.4 & \textbf{75.2} & 55.6 & 76.5 & 61.1 \\
            QueryInst\cite{article40} & ResNet-101 & 51.0 & 67.1 & 58.1 & 71.0 & 50.6 & 67.4 & 57.5 \\
            ISTR\cite{article37} & ResNet-101 & 62.7 & 80.8 & 70.8 & 73.2 & 62.0 & 80.7 & 70.2 \\
            Ours & ResNet-101 & \textbf{64.5} & \textbf{82.7} & \textbf{72.7} & 74.9 & \textbf{63.8} & \textbf{82.6} & \textbf{71.9} \\ \hline
        \end{tabular}}
    \end{table}
    \begin{table*}[!tbp]
        \vspace{-1cm}
        \caption{Performance comparisons on DocLayNet dataset.}
        \label{tab4}
        \resizebox{\textwidth}{!}{
        \begin{tabular}{c|c|ccccccccccc|c}
        \hline
        Method & Backbone & Caption & Footnote & Formula & List-item & Page-footer & Page-header & Picture & Section-header & Table & Text & Title & mAP \\ \hline
        Faster R-CNN\cite{article33} & R101 & 70.1 & 73.7 & 63.5 & 81.0 & 58.9 & 72.0 & 72.0 & 68.4 & 82.2 & 85.4 & 79.9 & 73.4 \\
        Mask R-CNN\cite{article35} & R50 & 68.4 & 70.9 & 60.1 & 81.2 & 61.6 & 71.9 & 71.7 & 67.6 & 82.2 & 84.6 & 76.7 & 72.4 \\
        Mask R-CNN\cite{article35} & R101 & 71.5 & 71.8 & 63.4 & 80.8 & 59.3 & 70.0 & 72.7 & 69.3 & 82.9 & 85.8 & 80.4 & 73.5 \\ 
        YOLOv5\cite{article48} & v5x6 & \textbf{77.7} & \textbf{77.2} & \textbf{66.2} & \textbf{86.2} & \textbf{61.1} & \textbf{67.9} & \textbf{77.1} & \textbf{74.6} & \textbf{86.3} & \textbf{88.1} & \textbf{82.7} & \textbf{76.8} \\ 
        Ours & R101 & 68.2 & 74.7 & 61.6 & 81.0 & 54.8 & 68.2 & 68.5 & 69.8 & 82.4 & 83.8 & 81.7 & 72.3 \\ \hline
        \end{tabular}}
    \end{table*}
    
    As shown in Table~\ref{tab3}, Mask R-CNN produced lower performance than Faster R-CNN. The same conclusion was reached for the DocLayNet dataset, as shown in Table~\ref{tab5}. It indicates that pixel-based image segmentation degrades performance when the dataset contains more complex document layouts. On the other hand, the recall rates of anchor-based methods are low. The reasons behind this include: (1) It is difficult to set an aspect ratio that can match all the instances. As shown in Figure~\ref{fig3} (a), the fourth and fifth columns present the results obtained using Mask R-CNN and our TransDLANet, both trained on our dataset. Even though we set eight anchor ratio scales, the experimental results show that Mask R-CNN still cannot correctly detect the advertisement instances (the bottom half of the newspaper page in the last row of column 4 of Figure~\ref{fig3} (a)) with large ratio scales but can only detect the paragraphs inside. (2) Anchor-based methods use non-maximum suppression to filter candidate bounding boxes. Therefore, if the overlapped area of the candidate bounding boxes of skewed neighboring document instances is large, they may be filtered out. This leads to detection errors and low recall.
 
    Our approach has achieved a remarkable mean average precision (mAP) of 64.5\% on the $M^{6}Doc$ dataset, surpassing the current state-of-the-art results. TransDLANet eliminates the need for complex anchor design by automatically learning to use a pre-set number of query vectors to encode and decode document instances in images. Additionally, the iterative refinement mechanism of TransDLANet helps overcome the challenges posed by dense arrangement, thereby reducing instance segmentation bias and achieving superior accuracy.
    
    \begin{table}[!htbp]
        \caption{Performance comparisons on PubLayNet dataset.}
        \label{tab5}
        \resizebox{\linewidth}{!}{
        \begin{tabular}{c|c|ccccc|c}
        \hline
        Method & Backbone & Text & Title & List & Table & Figure & mAP \\    \hline
        Faster R-CNN\cite{article33} & X101 & 91.0 & 82.6 & 88.3 & 95.4 & 93.7 & 90.2 \\
        Mask R-CNN\cite{article35} & X101 & 91.6 & 84.0 & 88.6 & 96.0 & 94.9 & 91.0 \\
        VSR~\cite{article18} & X101 & \textbf{96.7} & \textbf{93.1} & 94.7 & \textbf{97.4} & 96.4 & 95.7 \\
        Ours & R101 & 94.3 & 89.21 & \textbf{95.2} & 97.2 & \textbf{96.6} & 94.5 \\ \hline
        \end{tabular}}
    \end{table}
    \subsection{Performance of the TransDLANet in other datasets}
    We also conducted experiments on the existing layout dataset to explore the performance of TransDLANet. Tables \ref{tab4}, \ref{tab5} show the performance of our model on DocLayNet and PubLayNet. 

    Table~\ref{tab4} displays the performance of our model on the DocLayNet dataset. As evident from the results, our model's performance was comparatively lower than those of other models. Upon further investigation of the visualization results (available in the Supplementary Material), we identified the primary reason for the low accuracy as the fact that we set a fixed number of queries in advance. This design caused our model to miss some instances in the images when multiple queries corresponded to a single instance.
    
    Table~\ref{tab5} demonstrates that TransDLANet achieves comparable or even superior performances to the VSR model for AP in the list, table, and figure categories in the PubLayNet dataset. However, the performance in the text and title categories is inferior to that of VSR. This disparity could be attributed to the fact that VSR exploits both visual and semantic features. The text and title categories exhibit considerable differences in semantic features, so semantic branching can better recognize them. However, TransDLANet does not exploit this distinct feature, so performance is a bit lower compared to VSR.

    \subsection{Discussion of Failure Cases}
    The first row of Figure~\ref{fig3} (a) demonstrates the deficiency of both existing models and TransDLANet in detecting handwritten documents due to the unique characteristics of notes. Unlike published documents, handwritten notes are not standardized, and each person may use their own writing style, making them difficult to understand. In addition, the images and tables within the notes subset are not as visually prominent as in other documents, making detection even more challenging. Furthermore, the performance of the current model is unsatisfactory when dealing with real scenario files with significant distortion. Therefore, future researches can explore the use of document rectification~\cite{article55, article56} as a preliminary step ahead of current methods to solve this challenge. Whatmore, both the current models and the TransDLANet face difficulties in detecting instances that are either densely packed or skewed. Although TransDLANet tries to mitigate this problem by using a transformer encoder to learn the relevance of queries, the problem of missing instance objects still exists. We can solve this problem by training more epochs, but this model converges very slowly. Therefore, future research should further accelerate the convergence rate of TransDLANet and think about how to improve the model's recall.
     

    \section{Conclusion}
    In this paper, we introduce the new $M^{6}Doc$ dataset, consisting of seven subsets that were acquired using various methods, such as PDF to image conversion, document scanning, and photographing. To our knowledge, $M^{6}Doc$ is the first dataset that includes real-world scenario files, diverse formats, types, languages, layouts, and comprehensive definitions of logical labels. It can serve as a valuable benchmark for studying logical layout analysis, generic layout analysis, multi-modal layout analysis, formula identification, and table analysis.
    
    We carried out a comprehensive benchmark evaluation of $M^{6}Doc$ using multiple baselines and conducted detailed analyses. Our findings demonstrate the challenging nature of the $M^{6}Doc$ dataset and the effectiveness of the detailed label annotations.
    
    For future work, we aim to design specialized models based on the $M^{6}Doc$ dataset to address the issue of generic layout analysis. Additionally, we plan to explore the challenges of different languages for multi-modal models and consider how to unify visually and semantically consistent annotation formats. Furthermore, we aim to enhance the diversity of our dataset by including further document layouts and types, if possible, to enrich the layout and type diversity.

\subsubsection*{Acknowledgement}
This research is supported in part by NSFC (Grant No.: 61936003), Zhuhai Industry Core and Key Technology Research Project (no. 2220004002350), GD-NSF (No.2021A1515011870), and the Science and Technology Foundation of Guangzhou Huangpu Development District (Grant 2020GH17)

{\small
\bibliographystyle{ieee_fullname}
\bibliography{mybib}
}

\clearpage
\newpage
\appendix
\twocolumn[
    \section*{\centering\Large Supplementary Material}
]
\setcounter{figure}{0} 
\setcounter{table}{0} 

\section{Data annotation process}
Firstly, before the annotation process, we conducted a pre-annotation process. We provided 500 clean document images, 500 annotation samples, and a preliminary version of the annotation guidelines to 47 annotators. They performed labeling on the same images without discussion.

Secondly, we compared each annotator's annotation result with the samples and quantified the matching degree and errors. If there are significant differences in the labeling of the same document element by different annotators, we will research the relevant knowledge to ensure the most accurate label and update the annotation guideline.

Finally, in the final annotation process, all annotators performed annotation according to the final, refined version of the guideline (with clear annotation requirements for all confusing cases). When annotators encounter document elements in uncertain categories, we will provide timely feedback to ensure consistency in labeling. Following this way, in the final check process, the percentage of inconsistent annotations that we should resolve is within 5\%.

\section{Impact of Document Size}
\begin{figure}[!htbp]
    \centering
    \includegraphics[width=0.8\linewidth]{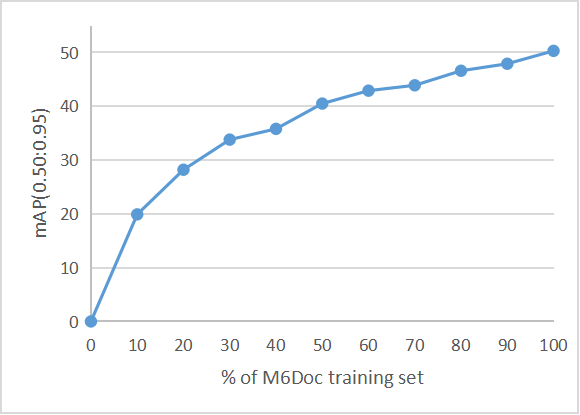}
    \caption{Mask R-CNN network with ResNet50 backbone trained on increasing fractions of the $M^{6}Doc$ dataset.}
    \label{fig4}
\end{figure}

To investigate the effect of dataset size on model performance, we trained Mask R-CNN on training sets of different proportions of $M^{6}Doc$ and evaluated the performance on a test set. As shown in Figure~\ref{fig4}, the mean accuracy (mAP) scores initially show rapid growth, and then slowly grow as the training dataset reaches sizes of 90\% and 100\%. This suggests that the performance of the model can still be improved if we continue to increase the size of the $M^{6}Doc$ dataset.

\vspace{-0.2cm}
\section{Dataset Comparison}
We conducted cross-validation experiments on the DocBank, PubLayNet, DocLayNet, and $M^{6}Doc$ datasets using the TransDLANetm model. A direct comparison is impossible due to the difference in label sets and annotation styles. Therefore, we focus on their common labels. We mapped the 74 labels in $M^{6}Doc$ to labels consistent with DocBank, PubLayNet, and DocLayNet, respectively. The mapping rules are: (1) excluding categories that are annotated with different methods. We exclude the List-item category because the consecutive lists are segmentally labeled in DocLayNet, whereas consecutive lists are combined into one object in DocBank, PubLayNet, and $M^{6}Doc$. (2) removing the categories that are specific to subsets. (3) Mapping the fine-grained labels to the coarse-grained labels of DocBank, DocLayNet, and PubLayNet. The mapping is shown in Tables~\ref{tab: mapping table M6Doc2DocBank}, ~\ref{tab: mapping table M6Doc2DocLayNet} and ~\ref{tab: mapping table M6Doc2PubLayNet}.

Tables ~\ref{tab: Dataset Comparison M6Doc2DocBank} and ~\ref{tab: Dataset Comparison M6Doc2PubLayNet} show that the models trained on $M^{6}Doc$, DocLayNet, and PubLayNet datasets achieve high accuracy on their own test sets. It is worth noting that the model trained on $M^{6}Doc$ maintained a high accuracy on the test sets of DocBank and PubLayNet, while the models trained on the DocBank and PubLayNet datasets performed inadequately on the test set of $M^{6}Doc$. This is due to the fact that DocBank and PubLayNet datasets have a single scenario and document type. It leads to poor scalability and robustness. In contrast, $M^{6}Doc$ contains a wide range of scenarios and includes document categories and scenarios from the other two datasets. It exhibits good robustness.

As shown in Table~\ref{tab: Dataset Comparison M6Doc2DocLayNet}, the model trained on DocLayNet, or $M^{6}Doc$ performs very well on their own test set, but has much lower performances on the foreign datasets. This is caused by the inconsistent layout of DocLayNet and the $M^{6}Doc$ datasets. Thus, it justifies the need for datasets with unseen layouts for the development document layout analysis.

\vspace{-0.2cm}
\section{Impact of Class Labels}
There are two versions of labeling the note subset, in which the first version (note-v1) contains 27 annotation categories and the second version (note-v2) contains 18 categories. Considering the fact that people take notes with large individual variability, the note subset presents higher ambiguity in annotations than other subsets. Therefore, we map the labels with ambiguity down to the paragraph category or delete them to obtain the second version. We use Mask R-CNN to train and evaluate the note subset with annotations from these two versions. As shown in Table~\ref{tab: note Class Labels}, after we reduce the ambiguity label of the model, the mAP is improved by 2.7\%.

\begin{table}[!htbp]
    \vspace{-1cm}
    \caption{A mapping table that maps the fine-grained labels of $M^{6}Doc$ to the coarse-grained labels of DocBank.}
    \label{tab: mapping table M6Doc2DocBank}
    \resizebox{\linewidth}{!}{
    \begin{tabular}{llll}
    \hline
    Before & After & Before & After \\ \hline
    QR code                     & -        & institute                    & -         \\
    advertisement               & figure   & jump line                    & -         \\
    algorithm                   & -        & kicker                       & -         \\
    answer                      & -        & lead                         & -         \\
    author                      & -        & marginal note                & -         \\
    barcode                     & -        & matching                     & -         \\
    bill                        & -        & mugshot                      & figure    \\
    blank                       & -        & option                       & -         \\
    bracket                     & -        & ordered list                 & -         \\
    breakout                    & -        & other question number        & -         \\
    byline                      & -        & page number                  & -         \\
    caption                     & caption  & paragraph                    & -         \\
    catalogue                   & -        & part                         & section   \\
    chapter title               & title    & play                         & -         \\
    code                        & -        & poem                         & -         \\
    correction                  & -        & reference                    & reference \\
    credit                      & -        & sealing line                 & -         \\
    dateline                    & -        & second-level question number & -         \\
    drop cap                    & -        & second-level title           & section   \\
    editor's note               & -        & section                      & -         \\
    endnote                     & -        & section title                & section   \\
    examinee information        & -        & sidebar                      & -         \\
    fifth-level title           & section  & sub section title            & section   \\
    figure                      & figure   & subhead                      & section   \\
    first-level question number & -        & subsub section title         & section   \\
    first-level title           & section  & supplementary note           & -         \\
    flag                        & -        & table                        & table     \\
    folio                       & -        & table caption                & caption   \\
    footer                      & -        & table note                   & -         \\
    footnote                    & -        & teasers                      & -         \\
    formula                     & equation & third-level question number  & -         \\
    fourth-level section title  & section  & third-level title            & section   \\
    fourth-level title          & section  & title                        & title     \\
    header                      & -        & translator                   & -         \\
    headline                    & section  & underscore                   & -         \\
    index                       & -        & unordered list               & -         \\
    inside                      & -        & weather forecast             & -        \\ \hline
    \end{tabular}}
\end{table}

\begin{table}[!htbp]
    \vspace{-0.2cm}
    \caption{A mapping table that maps the fine-grained labels of $M^{6}Doc$ to the coarse-grained labels of DocLayNet.}
    \label{tab: mapping table M6Doc2DocLayNet}
    \resizebox{\linewidth}{!}{
    \begin{tabular}{llll}
    \hline
    Before & After & Before & After \\ \hline
    QR code & - & institute & Text \\
    advertisement & Picture & jump line & Text \\
    algorithm & - & kicker & Text \\
    answer & - & lead & Text \\
    author & Text & marginal note & Page-header \\
    barcode & - & matching & - \\
    bill & - & mugshot & Picture \\
    blank & - & option & - \\
    bracket & - & ordered list & - \\
    breakout & Text & other question number & - \\
    byline & Text & page number & Text \\
    caption & Caption & paragraph & Text \\
    catalogue & - & part & Title \\
    chapter title & Title & play & - \\
    code & - & poem & - \\
    correction & - & reference & - \\
    credit & Text & sealing line & - \\
    dateline & Text & second-level question number & - \\
    drop cap & - & second-level title & Title \\
    editor's note & Text & section & Text \\
    endnote & Text & section title & Title \\
    examinee information & - & sidebar & - \\
    fifth-level title & Title & sub section title & Title \\
    figure & Picture & subhead & Title \\
    first-level question number & - & subsub section title & Title \\
    first-level title & Title & supplementary note & - \\
    flag & - & table & Table \\
    folio & Section-header & table caption & Caption \\
    footer & Page-footer & table note & - \\
    footnote & Footnote & teasers & - \\
    formula & Formula & third-level question number & - \\
    fourth-level section title & Title & third-level title & Title \\
    fourth-level title & Title & title & Title \\
    header & Section-header & translator & Text \\
    headline & Title & underscore & - \\
    index & Page-header & unordered list & - \\
    inside & - & weather forecast & - \\ \hline
    \end{tabular}}
    \vspace{-0.5cm}
\end{table}

\begin{table}[!htbp]
    \vspace{-1cm}
    \caption{A mapping table that maps the fine-grained labels of $M^{6}Doc$ to the coarse-grained labels of PubLayNet.}
    \label{tab: mapping table M6Doc2PubLayNet}
    \resizebox{\linewidth}{!}{
    \begin{tabular}{llll}
    \hline
    Before & After & Before & After \\ \hline
    QR code                     & -      & institute                    & text   \\
    advertisement               & figure & jump line                    & text   \\
    algorithm                   & -      & kicker                       & text   \\
    answer                      & text   & lead                         & text   \\
    author                      & text   & marginal note                & text   \\
    barcode                     & -      & matching                     & -      \\
    bill                        & -      & mugshot                      & figure \\
    blank                       & -      & option                       & -      \\
    bracket                     & -      & ordered list                 & list   \\
    breakout                    & text   & other question number        & -      \\
    byline                      & text   & page number                  & text   \\
    caption                     & text   & paragraph                    & text   \\
    catalogue                   & -      & part                         & title  \\
    chapter title               & title  & play                         & -      \\
    code                        & -      & poem                         & -      \\
    correction                  & -      & reference                    & -      \\
    credit                      & text   & sealing line                 & -      \\
    dateline                    & text   & second-level question number & -      \\
    drop cap                    & -      & second-level title           & Title  \\
    editor's note               & text   & section                      & text   \\
    endnote                     & text   & section title                & title  \\
    examinee information        & -      & sidebar                      & -      \\
    fifth-level title           & title  & sub section title            & title  \\
    figure                      & figure & subhead                      & title  \\
    first-level question number & -      & subsub section title         & title  \\
    first-level title           & title  & supplementary note           & -      \\
    flag                        & -      & table                        & table  \\
    folio                       & text   & table caption                & text   \\
    footer                      & text   & table note                   & text   \\
    footnote                    & text   & teasers                      & -      \\
    formula                     & text   & third-level question number  & -      \\
    fourth-level section title  & title  & third-level title            & title  \\
    fourth-level title          & title  & title                        & title  \\
    header                      & text   & translator                   & text   \\
    headline                    & text   & underscore                   & -      \\
    index                       & text   & unordered list               & list   \\
    inside                      & -      & weather forecast             & -     \\ \hline
    \end{tabular}}
\end{table}

\begin{table}[!htbp]
    \vspace{-0.2cm}
    \caption{The prediction performance (mAP@0.5-0.95) of the TransDLANet network was evaluated on the common label classes of the DocBank and $M^{6}Doc$ datasets.}
    \label{tab: Dataset Comparison M6Doc2DocBank}
    \centering
    \resizebox{0.65\linewidth}{!}{
    \begin{tabular}{llcc}
    \hline
    \multicolumn{1}{c}{\multirow{2}{*}{Training on}} & \multicolumn{1}{c}{\multirow{2}{*}{labels}} & \multicolumn{2}{c}{Testing on} \\ 
    \multicolumn{1}{c}{} & \multicolumn{1}{c}{} & $M^{6}Doc$ & DocBank \\ \hline
    \multirow{4}{*}{$M^{6}Doc$}   & figure & 69.77 & 42.67 \\
                             & table  & 72.57 & 43.29 \\
                             & title  & 58.16 & 36.47 \\
                             & mAP    & 66.83 & 40.81 \\ \hline
    \multirow{4}{*}{DocBank} & figure & 20.70 & 58.47 \\
                             & table  & 18.01 & 62.98 \\
                             & title  & 7.26  & 83.70 \\
                             & mAP    & 15.32 & 68.38 \\ \hline
    \end{tabular}}
    \vspace{-0.2cm}
\end{table}

\begin{table}[!htbp]
    \vspace{-1cm}
    \caption{The prediction performance (mAP@0.5-0.95) of the TransDLANet network was evaluated on the common label classes of the PubLayNet and $M^{6}Doc$ datasets.}
    \label{tab: Dataset Comparison M6Doc2PubLayNet}
    \centering
    \resizebox{0.65\linewidth}{!}{
    \begin{tabular}{llcc}
    \hline
    \multicolumn{1}{c}{\multirow{2}{*}{Training on}} & \multicolumn{1}{c}{\multirow{2}{*}{labels}} & \multicolumn{2}{c}{Testing on} \\ 
    \multicolumn{1}{c}{} & \multicolumn{1}{c}{} & $M^{6}Doc$ & PubLayNet \\ \hline
    \multirow{6}{*}{$M^{6}Doc$}     & Text   & 72.56 & 60.21 \\
                               & Title  & 63.50 & 53.26 \\
                               & List   & 38.95 & 59.15 \\
                               & Table  & 74.83 & 79.66 \\
                               & Figure & 74.23 & 62.45 \\
                               & mAP    & 64.81 & 62.94 \\ \hline
    \multirow{6}{*}{PubLayNet} & Text   & 20.46 & 94.26 \\
                               & Title  & 12.92 & 89.20 \\
                               & List   & 7.41  & 95.18 \\
                               & Table  & 12.98 & 97.21 \\
                               & Figure & 8.39  & 96.62 \\
                               & mAP    & 12.43 & 94.49 \\ \hline
    \end{tabular}}
\end{table}

\begin{table}[!htbp]
    \caption{The prediction performance (mAP@0.5-0.95) of the TransDLANet network was evaluated on the common label classes of the DocLayNet and $M^{6}Doc$ datasets.}
    \label{tab: Dataset Comparison M6Doc2DocLayNet}
    \centering
    \resizebox{0.85\linewidth}{!}{
    \begin{tabular}{llcc}
    \hline
    \multicolumn{1}{c}{\multirow{2}{*}{Training on}} & \multicolumn{1}{c}{\multirow{2}{*}{labels}} & \multicolumn{2}{c}{Testing on} \\ 
    \multicolumn{1}{c}{} & \multicolumn{1}{c}{} & $M^{6}Doc$ & DocLayNet \\ \hline
     & Caption & 61.9 & 12.7 \\
     & Footnote & 70.2 & 5.8 \\
     & Formula & 47.7 & 9.0 \\
     & Page-footer & 71.0 & 8.0 \\
     $M^{6}Doc$ & Page-header & 71.1 & 3.2 \\
     & Picture & 75.4 & 30.0 \\
     & Section-header & 73.2 & 5.0 \\
     & Table & 78.0 & 34.2 \\
     & Text & 80.0 & 26.2 \\
     & Title & 71.1 & 0.4 \\ 
     & mAP & 69.96 & 13.45 \\ \hline
     & Caption & 13.2 & 68.2 \\ 
     & Footnote & 7.0 & 74.7 \\
     & Formula & 2.5 & 61.6 \\
     & Page-footer & 8.2 & 54.8 \\
     DocLayNet & Page-header & 0.8 & 68.2 \\
     & Picture & 40.1 & 68.5 \\
     & Section-header & 1.6 & 69.8 \\
     & Table & 39.2 & 82.4 \\
     & Text & 45.8 & 83.8 \\
     & Title & 3.6 & 81.7 \\
     & mAP & 16.20 & 71.37 \\ \hline
    \end{tabular}}
\end{table}

\begin{table}[!htbp]
    \vspace{-1cm}
    \centering
    \caption{Effects of coarse and fine granularity of labels on the note dataset.}
    \label{tab: note Class Labels}
    \renewcommand{\arraystretch}{1}
    \resizebox{0.73\linewidth}{!}{
    \begin{tabular}{lcc}
    \hline
    \textbf{Category} & \textbf{note\_v1} & \textbf{note\_v2} \\
    \hline
    answer & 8.1 & 5.8 \\
    bracket & 0.0 & - \\
    caption & 0.0 & 0.1 \\
    catalogue & 19.2 & 14.3 \\
    chapter title & 18.0 & 18.3 \\
    fifth-level title & 2.4 & paragraph \\
    figure & 0.4 & 0.7 \\
    first-level question number & 0.0 & - \\
    first-level title & 13.6 & paragraph \\
    footer & 62.5 & 58.5 \\
    formula & 1.5 & 2.6 \\
    fourth-level title & 19.5 & paragraph \\
    option & 0.0 & 0 \\
    ordered list & 3.2 & 2.2 \\
    page number & 55.3 & 55.3 \\
    paragraph & 28.1 & 41.3 \\
    part & 0.0 & 0 \\
    second-level question number & 0.0 & - \\
    second-level title & 0.0 & paragraph \\
    section & 12.4 & 17 \\
    section title & 9.3 & 7 \\
    sub section title & 5.1 & 5.9 \\
    supplementary note & 0.0 & 0 \\
    table & 22.7 & 17.4 \\
    third-level title & 25.8 & paragraph \\
    underscore & 0.0 & - \\ 
    unordered list & 28.5 & 25.9 \\ 
    mAP & 12.4 & 15.1 \\ \hline
    \end{tabular}}
    \end{table}

\begin{table}[!htbp]
    \centering
    \caption{Ablation study for mask embedding dimension.}
    \label{tab: mask embedding}
    \resizebox{\linewidth}{!}{
    \begin{tabular}{lccccccc}
    \hline
    \multirow{3}{*}{\textbf{Ablation study}} & \multicolumn{4}{c}{Object} & \multicolumn{3}{c}{Instance} \\
    & \multicolumn{4}{c}{Detection} & \multicolumn{3}{c}{Segmentation} \\ \cline{2-8}
    & mAP & AP50 & AP75 & Recall & mAP  & AP50 & AP75 \\ \hline
    embedding dimension = 20 & 63.2 & 81.0 & 72.0 & 74.0 & 62.7 & 80.9 & 71.3 \\ 
    embedding dimension = 40 & \textbf{64.5} & \textbf{82.7} & \textbf{72.7} & \textbf{74.9} & \textbf{63.8} & \textbf{82.6} & \textbf{71.9}  \\ 
    embedding dimension = 60 & 63.4 & 81.1 & 74.6 & 72.3 & 62.8 & 81.0 & 71.3 \\ \hline 
    \end{tabular}}
\end{table}
\begin{table}[!htbp]
    \centering
    \caption{Ablation study for different components.}
    \label{tab: different components}
    \resizebox{\linewidth}{!}{
    \begin{tabular}{lccccccc}
    \hline
    \multirow{3}{*}{\textbf{Ablation study}} & \multicolumn{4}{c}{Object} & \multicolumn{3}{c}{Instance} \\
    & \multicolumn{4}{c}{Detection} & \multicolumn{3}{c}{Segmentation} \\ \cline{2-8}
    & mAP & AP50 & AP75 & Recall & mAP & AP50 & AP75  \\ \hline
    Ours, w/o Transformer encoder & 47.8 & 62.6 & 54.4 & 65.4 & 47.3 & 62.6 & 53.9  \\ 
    Ours, w/o Dynamic decoder & 52.8 & 70.5 & 48.0 & 73.9 & 52.3 & 70.4 & 47.6  \\ 
    Ours, w/o Shared\_MLP & 64.2 & 82.3 & 72.1 & 74.1 & 63.6 & 82.2 & 71.2  \\ 
    Ours & \textbf{64.5} & \textbf{82.7} & \textbf{72.7} & \textbf{74.9} & \textbf{63.8} & \textbf{82.6} & \textbf{71.9}  \\ \hline
    \end{tabular}}
\end{table}

\section{Ablation study for TransDLANet}
\textbf{Mask Embedding}. Table~\ref{tab: mask embedding} shows the results of mask embedding in different dimensions. Because the profile of document elements is relatively simple, a mask dimension setting of 40 can obtain the best performance.

\textbf{Transformer encoder}. The biggest limitation of the query-based approach is its low recall. In order to improve the recall, we differ from the ISTR approach in that we first use a standard Transformer encoder, which performs self-attentive feature learning on the implicit embedding vectors of the query vector and uses an adaptive element matching mechanism to enhance the association between document instances encoded by the query vector. As shown in Table~\ref{tab: different components}, we conducted experiments to test the effectiveness of the Transformer encoder. Compared with the methods without a Transformer encoder, there is a performance gap of around 20\%.

\textbf{Dynamic decoder}. After we get the information between queries in the Transformer encoder, we went through the Dynamic decoder to fuse the RoI and image features. We also conducted experiments to test the effectiveness of the Dynamic encoder. As shown in Table~\ref{tab: different components}, compared with using the fusion feature, there is an improvement of around 10\%.

\textbf{Shared MLP}. We use three shared-parameter MLP branches to decode the fused interaction features for multi-task learning. Compared with the methods without shared-parameter MLP, there is an improvement of around 0.3-0.8\%.

\begin{table*}[!htbp]
\caption{Performance comparisons on nine subsets of $M^{6}Doc$.}
\label{tab: nine subsets results}
\resizebox{\textwidth}{!}{
\begin{tabular}{l|l|ccc|ccc|ccc|ccc|ccc|ccc}
    \hline
    \multicolumn{1}{c|}{} & \multicolumn{1}{c|}{} & \multicolumn{6}{c|}{\textbf{scientific article}} & \multicolumn{6}{c|}{\textbf{magazine\_ch}} & \multicolumn{6}{c}{\textbf{magazine\_en}} \\ \cline{3-20}
    \multicolumn{1}{c|}{} & \multicolumn{1}{c|}{} & \multicolumn{3}{c|}{Object} & \multicolumn{3}{c|}{Instance} & \multicolumn{3}{c|}{Object} & \multicolumn{3}{c|}{Instance} & \multicolumn{3}{c|}{Object} & \multicolumn{3}{c}{Instance} \\ 
    \multicolumn{1}{c|}{} & \multicolumn{1}{c|}{} & \multicolumn{3}{c|}{Detection} & \multicolumn{3}{c|}{Segmentation} & \multicolumn{3}{c|}{Detection} & \multicolumn{3}{c|}{Segmentation} & \multicolumn{3}{c|}{Detection} & \multicolumn{3}{c}{Segmentation} \\ \cline{3-20}
    \multicolumn{1}{c|}{\multirow{-4}{*}{Method}} & \multicolumn{1}{c|}{\multirow{-4}{*}{Backbone}} & mAP & AP50 & AP75 & mAP & AP50 & AP75 & mAP & AP50 & AP75 & mAP & AP50 & AP75 & mAP & AP50 & AP75 & mAP & AP50 & AP75 \\ \hline
    FCOS & ResNet-101 & 26.3 & 45.1 & 27.2 & 25.9 & 44.9 & 26.5 & 40.1 & 57.3 & 45.8 & 39.7 & 57.3 & 45.1 & 38.4 & 60.5 & 42.5 & 37.8 & 60.5 & 41.8 \\
    FoveaBox & ResNet-101 & 29.8 & 52.7 & 30.7 & 29.4 & 52.4 & 30.8 & 43.4 & 59.7 & 50.4 & 43.1 & 59.7 & 50.0 & 41.5 & 66.7 & 44.0 & 41.1 & 66.9 & 42.8 \\
    Faster R-CNN & ResNet-101 & 41.5 & 62.0 & 46.8 & 40.9 & 61.7 & 45.4 & 49.0 & 63.5 & 58.3 & 48.4 & 63.5 & 57.1 & 47.9 & 66.7 & 55.9 & 47.1 & 66.7 & 54.5 \\
    Cascade R-CNN & ResNet-101 & 39.8 & 55.5 & 45.7 & 39.4 & 55.8 & 44.8 & 51.3 & 63.5 & 60.0 & 50.7 & 63.4 & 59.5 & 46.3 & 61.3 & 54.0 & 45.9 & 61.2 & 53.3 \\
    Mask R-CNN & ResNet-101 & 34.9 & 53.5 & 37.6 & 35.0 & 53.3 & 38.3 & 47.1 & 61.1 & 55.4 & 46.4 & 61.0 & 55.2 & 43.9 & 60.8 & 50.2 & 43.2 & 60.7 & 50.3 \\
    Cascade Mask R-CNN & ResNet-101 & 41.8 & 57.3 & 47.4 & 41.4 & 57.1 & 46.6 & 46.4 & 58.6 & 54.0 & 46.0 & 58.6 & 53.9 & 59.4 & 74.7 & 69.0 & 58.5 & 74.9 & 68.3 \\
    HTC & ResNet-101 & 49.2 & 66.0 & 55.2 & 48.8 & 65.9 & 54.3 & \textbf{51.9} & \textbf{64.7} & \textbf{60.3} & \textbf{50.7} & \textbf{64.8} & \textbf{59.4} & 60.4 & 77.3 & 71.7 & 59.6 & 77.3 & 70.9 \\
    SCNet & ResNet-101 & 36.0 & 51.4 & 40.9 & 35.5 & 51.3 & 39.4 & 49.0 & 62.2 & 57.2 & 48.3 & 62.2 & 56.9 & 49.1 & 66.3 & 57.3 & 48.2 & 66.2 & 56.2 \\
    SOLO & ResNet-101 & 32.1 & 51.1 & 33.9 & 32.8 & 53.5 & 33.9 & 35.6 & 53.0 & 39.9 & 37.1 & 54.6 & 42.1 & 34.4 & 59.9 & 32.8 & 36.1 & 59.6 & 34.5 \\
    SOLOv2 & ResNet-101 & 33.5 & 54.0 & 35.9 & 33.0 & 54.5 & 36.0 & 33.8 & 51.8 & 36.5 & 35.8 & 53.7 & 39.5 & 45.3 & 71.1 & 49.4 & 47.9 & 72.7 & 54.0 \\
    Deformable DETR & ResNet-101 & 32.3 & 43.7 & 35.8 & 32.0 & 43.7 & 35.3 & 40.2 & 55.1 & 45.8 & 39.9 & 55.0 & 45.0 & 51.1 & 72.0 & 58.6 & 50.8 & 71.9 & 57.7 \\
    QueryInst & ResNet-101 & 32.0 & 46.2 & 36.3 & 31.6 & 45.8 & 35.5 & 37.4 & 49.7 & 43.2 & 37.6 & 50.4 & 43.5 & 44.8 & 60.6 & 53.8 & 44.5 & 61.1 & 53.2 \\
    ISTR & ResNet-101 & \textbf{61.8} & \textbf{80.3} & \textbf{69.7} & \textbf{61.1} & \textbf{80.2} & \textbf{70.2} & 50.5 & 63.4 & 58.4 & 50.5 & 63.5 & 58.4 & 66.3 & 83.0 & 75.6 & 65.7 & 83.0 & 75.0 \\
    Ours & ResNet-101 & 59.7 & 78.7 & 68.2 & 59.1 & 78.5 & 67.0 & 50.2 & 63.0 & 57.7 & 49.8 & 62.9 & 57.3 & \textbf{68.2} & \textbf{85.0} & \textbf{77.2} & \textbf{68.2} & \textbf{85.0} & \textbf{77.2} \\ \hline
    \multicolumn{1}{c|}{} & \multicolumn{1}{c|}{} & \multicolumn{6}{c|}{\textbf{note}} & \multicolumn{6}{c|}{\textbf{newspaper\_ch}} & \multicolumn{6}{c}{\textbf{newspaper\_en}} \\  \cline{3-20}
    \multicolumn{1}{c|}{} & \multicolumn{1}{c|}{} & \multicolumn{3}{c|}{Object} & \multicolumn{3}{c|}{Instance} & \multicolumn{3}{c|}{Object} & \multicolumn{3}{c|}{Instance} & \multicolumn{3}{c|}{Object} & \multicolumn{3}{c}{Instance} \\
    \multicolumn{1}{c|}{} & \multicolumn{1}{c|}{} & \multicolumn{3}{c|}{Detection} & \multicolumn{3}{c|}{Segmentation} & \multicolumn{3}{c|}{Detection} & \multicolumn{3}{c|}{Segmentation} & \multicolumn{3}{c|}{Detection} & \multicolumn{3}{c}{Segmentation} \\ \cline{3-20}
    \multicolumn{1}{c|}{\multirow{-4}{*}{Method}} & \multicolumn{1}{c|}{\multirow{-4}{*}{Backbone}} & mAP & AP50 & AP75 & mAP & AP50 & AP75 & mAP & AP50 & AP75 & mAP & AP50 & AP75 & mAP & AP50 & AP75 & mAP & AP50 & AP75 \\ \hline
    GFL & ResNet-101 & 11.1 & 19.1 & 11.7 & 11.0 & 19.1 & 12.1 & 22.1 & 35.9 & 24.7 & 21.8 & 35.9 & 23.9 & 20.5 & 30.1 & 22.8 & 20.3 & 30.0 & 22.7 \\
    FCOS & ResNet-101 & 19.1 & 36.7 & 18.7 & 18.9 & 36.5 & 17.8 & 22.7 & 41.7 & 21.1 & 22.5 & 41.6 & 21.7 & 17.8 & 32.6 & 17.4 & 17.5 & 32.4 & 16.8 \\
    FoveaBox & ResNet-101 & 19.8 & 36.2 & 21.5 & 19.5 & 36.3 & 20.3 & 21.5 & 37.6 & 22.3 & 21.3 & 37.5 & 22.1 & 35.5 & 56.4 & 39.6 & 34.9 & 56.3 & 37.7 \\
    Faster R-CNN & ResNet-101 & 29.3 & 46.1 & 33.9 & 28.9 & 46.1 & 32.9 & 32.2 & 50.6 & 33.9 & 32.3 & 50.8 & 33.9 & 34.3 & 50.7 & 39.4 & 34.0 & 50.7 & 39.3 \\
    Cascade R-CNN & ResNet-101 & 22.5 & 34.9 & 27.3 & 22.3 & 34.9 & 27.3 & 27.7 & 42.6 & 29.8 & 27.7 & 42.7 & 30.0 & 26.3 & 36.5 & 29.6 & 26.0 & 36.3 & 29.3 \\
    Mask R-CNN & ResNet-101 & 15.1 & 27.6 & 15.3 & 15.2 & 27.8 & 14.6 & 21.2 & 36.9 & 21.2 & 20.5 & 36.2 & 19.9 & 19.9 & 31.1 & 21.9 & 19.7 & 31.0 & 21.8 \\
    Cascade Mask R-CNN & ResNet-101 & 24.3 & 36.7 & 28.9 & 24.0 & 36.7 & 28.0 & 43.2 & 60.8 & 47.1 & 42.9 & 60.7 & 47.1 & 23.4 & 32.8 & 26.9 & 23.1 & 32.7 & 26.6 \\
    HTC & ResNet-101 & 36.7 & 53.4 & 43.4 & 36.7 & 53.7 & 42.3 & 36.3 & 53.3 & 38.8 & 5.6 & 53.1 & 37.9 & 43.7 & 57.3 & 48.4 & 43.4 & 57.1 & 48.1 \\
    SCNet & ResNet-101 & 27.9 & 41.9 & 33.8 & 27.9 & 41.6 & 33.0 & 20.0 & 33.0 & 20.7 & 19.9 & 32.8 & 20.7 & 19.3 & 27.2 & 22.3 & 19.2 & 27.2 & 21.9 \\
    SOLO & ResNet-101 & 22.2 & 38.0 & 22.8 & 22.1 & 39.3 & 23.8 & 30.5 & 48.0 & 33.1 & 30.9 & 48.5 & 34.2 & 14.5 & 32.7 & 11.6 & 14.9 & 31.8 & 14.1 \\
    SOLOv2 & ResNet-101 & 26.9 & 44.1 & 28.5 & 29.0 & 44.7 & 32.7 & 24.5 & 40.2 & 26.1 & 26.2 & 42.5 & 28.0 & 30.7 & 50.1 & 31.5 & 32.7 & 51.8 & 34.9 \\
    Deformable DETR & ResNet-101 & 24.2 & 33.5 & 28.5 & 23.9 & 33.5 & 28.3 & 29.7 & 43.8 & 32.4 & 29.6 & 43.7 & 32.3 & 34.2 & 49.7 & 38.1 & 34.1 & 49.3 & 38.4 \\
    QueryInst & ResNet-101 & 23.3 & 35.5 & 27.1 & 23.3 & 35.5 & 26.5 & 28.9 & 43.6 & 31.5 & 29.3 & 44.7 & 31.8 & 36.5 & 48.2 & 41.4 & 38.4 & 51.2 & 43.4 \\
    ISTR & ResNet-101 & \textbf{48.6} & \textbf{63.9} & \textbf{57.3} & \textbf{48.5} & \textbf{63.9} & \textbf{56.7} & 52.7 & 68.3 & 58.9 & 52.3 & 68.4 & 58.0 & 61.0 & 73.9 & 68.8 & 60.7 & 73.9 & 68.1 \\ 
    Ours & ResNet-101 & 44.1 & 60.5 & 50.7 & 43.6 & 60.3 & 49.9 & \textbf{59.4} & \textbf{78.1} & \textbf{65.9} & \textbf{59.0} & \textbf{78.1} & \textbf{65.3} & \textbf{64.0} & \textbf{78.4} & \textbf{73.3} & \textbf{63.6} & \textbf{78.2} & \textbf{72.6} \\ \hline
    \multicolumn{1}{c|}{} & \multicolumn{1}{c|}{} & \multicolumn{6}{c|}{\textbf{test paper}} & \multicolumn{6}{c|}{\textbf{textbook}} & \multicolumn{6}{c}{\textbf{book}} \\ \cline{3-20}
    \multicolumn{1}{c|}{} & \multicolumn{1}{c|}{} & \multicolumn{3}{c|}{Object} & \multicolumn{3}{c|}{Instance} & \multicolumn{3}{c|}{Object} & \multicolumn{3}{c|}{Instance} & \multicolumn{3}{c|}{Object} & \multicolumn{3}{c}{Instance} \\
    \multicolumn{1}{c|}{} & \multicolumn{1}{c|}{} & \multicolumn{3}{c|}{Detection} & \multicolumn{3}{c|}{Segmentation} & \multicolumn{3}{c|}{Detection} & \multicolumn{3}{c|}{Segmentation} & \multicolumn{3}{c|}{Detection} & \multicolumn{3}{c}{Segmentation} \\ \cline{3-20}
    \multicolumn{1}{c|}{\multirow{-4}{*}{Method}} & \multicolumn{1}{c|}{\multirow{-4}{*}{Backbone}} & mAP & AP50 & AP75 & mAP & AP50 & AP75 & mAP & AP50 & AP75 & mAP & AP50 & AP75 & mAP & AP50 & AP75 & mAP & AP50 & AP75 \\ \hline
    GFL & ResNet-101 & 44.4 & 65.4 & 52.1 & 43.4 & 65.1 & 49.6 & 38.7 & 57.1 & 44.9 & 37.1 & 57.0 & 42.5 & 8.0 & 15.8 & 6.8 & 6.8 & 15.5 & 4.7 \\
    FCOS & ResNet-101 & 37.9 & 59.7 & 43.0 & 36.6 & 59.6 & 39.4 & 33.2 & 52.7 & 38.0 & 31.6 & 52.5 & 34.8 & 10.3 & 23.5 & 6.7 & 7.9 & 20.7 & 4.2 \\
    FoveaBox & ResNet-101 & 42.8 & 66.8 & 48.9 & 41.4 & 66.8 & 46.3 & 34.5 & 54.2 & 39.7 & 33.2 & 54.1 & 37.3 & 21.6 & 37.6 & 22.3 & 21.3 & 37.5 & 22.1 \\
    Faster R-CNN & ResNet-101 & 52.0 & 74.1 & 60.6 & 51.1 & 74.2 & 59.8 & 43.6 & 62.7 & 52.7 & 42.4 & 62.6 & 50.2 & 14.5 & 27.2 & 13.5 & 12.0 & 26.3 & 9.2 \\
    Cascade R-CNN & ResNet-101 & 54.3 & 73.4 & 63.0 & 53.8 & 73.5 & 62.7 & 47.1 & 64.7 & 55.4 & 45.8 & 64.6 & 54.2 & 9.2 & 16.3 & 8.9 & 7.7 & 16.2 & 6.0 \\
    Mask R-CNN & ResNet-101 & 49.1 & 70.2 & 56.9 & 48.1 & 70.0 & 55.7 & 40.3 & 58.1 & 48.3 & 39.8 & 58.0 & 47.5 & 7.1 & 14.9 & 5.7 & 7.6 & 15.0 & 5.9 \\
    Cascade Mask R-CNN & ResNet-101 & 52.6 & 71.7 & 61.1 & 52.3 & 71.8 & 60.4 & 45.7 & 62.8 & 54.1 & 44.4 & 62.7 & 52.3 & 10.8 & 18.8 & 11.5 & 8.8 & 18.4 & 7.1 \\
    HTC & ResNet-101 & 57.9 & 77.7 & 66.9 & 57.2 & 77.6 & 65.9 & 51.2 & 69.6 & 60.3 & 50.5 & 69.6 & 59.1 & 19.6 & 29.6 & 24.1 & 18.8 & 29.5 & 22.4 \\
    SCNet & ResNet-101 & 54.8 & 75.5 & 64.2 & 53.9 & 75.3 & 62.6 & 44.6 & 62.7 & 51.8 & 43.6 & 62.5 & 50.3 & 6.6 & 12.2 & 6.3 & 6.7 & 12.2 & 6.3 \\
    SOLO & ResNet-101 & 36.2 & 59.1 & 38.5 & 36.2 & 61.1 & 37.9 & 31.8 & 49.7 & 35.1 & 31.1 & 50.3 & 34.8 & 5.8 & 13.9 & 4.0 & 3.2 & 10.0 & 1.1 \\
    SOLOv2 & ResNet-101 & 33.0 & 55.7 & 33.3 & 34.5 & 57.4 & 36.5 & 33.7 & 54.6 & 36.6 & 35.0 & 55.3 & 38.2 & 17.3 & 29.8 & 17.2 & 15.6 & 29.2 & 16.4 \\
    Deformable DETR & ResNet-101 & 53.7 & 75.2 & 60.8 & 53.5 & 75.3 & 60.2 & 46.6 & 64.6 & 53.8 & 45.0 & 64.4 & 51.5 & 14.0 & 21.7 & 15.5 & 10.1 & 20.0 & 9.0 \\
    QueryInst & ResNet-101 & 44.0 & 60.9 & 50.4 & 43.4 & 61.2 & 49.9 & 35.7 & 50.0 & 41.3 & 35.5 & 50.1 & 41.1 & 10.7 & 17.1 & 11.8 & 10.5 & 17.2 & 11.6 \\
    ISTR & ResNet-101 & 60.4 & 80.9 & \textbf{68.5} & 60.1 & 80.9 & \textbf{67.9} & 50.1 & 68.2 & 58.7 & 49.5 & 68.1 & 57.3 & \textbf{29.0} & 39.9 & \textbf{35.4} & \textbf{28.4} & 39.8 & \textbf{34.4} \\
    Ours & ResNet-101 & \textbf{60.7} & \textbf{81.9} & 68.0 & \textbf{60.3} & \textbf{82.2} & 66.9 & \textbf{51.7} & \textbf{70.1} & \textbf{60.3} & \textbf{51.2} & \textbf{70.1} & \textbf{59.5} & 28.3 & \textbf{41.0} & 33.0 & 27.9 & \textbf{40.7} & 33.0 \\ \hline
\end{tabular}}
\end{table*}

\section{Performance of baseline and TransDLANet on the nine subsets of $M^{6}Doc$}


We trained TransDLANet for 500 epochs, and the learning rate was reduced to $2 \times 10^{-6}$ and $2 \times 10^{-7}$ at 50\% and 75\% of the training epochs, respectively. We use object detection-based, instance segmentation-based, and query-based approaches to experiment with nine subsets of $M^{6}Doc$.

As shown in Table~\ref{tab: nine subsets results}, we can see that TransDLANet obtains the best performance in almost all subsets.

\begin{figure*}[!htbp]
    \centering
    \includegraphics[width=0.85\textwidth]{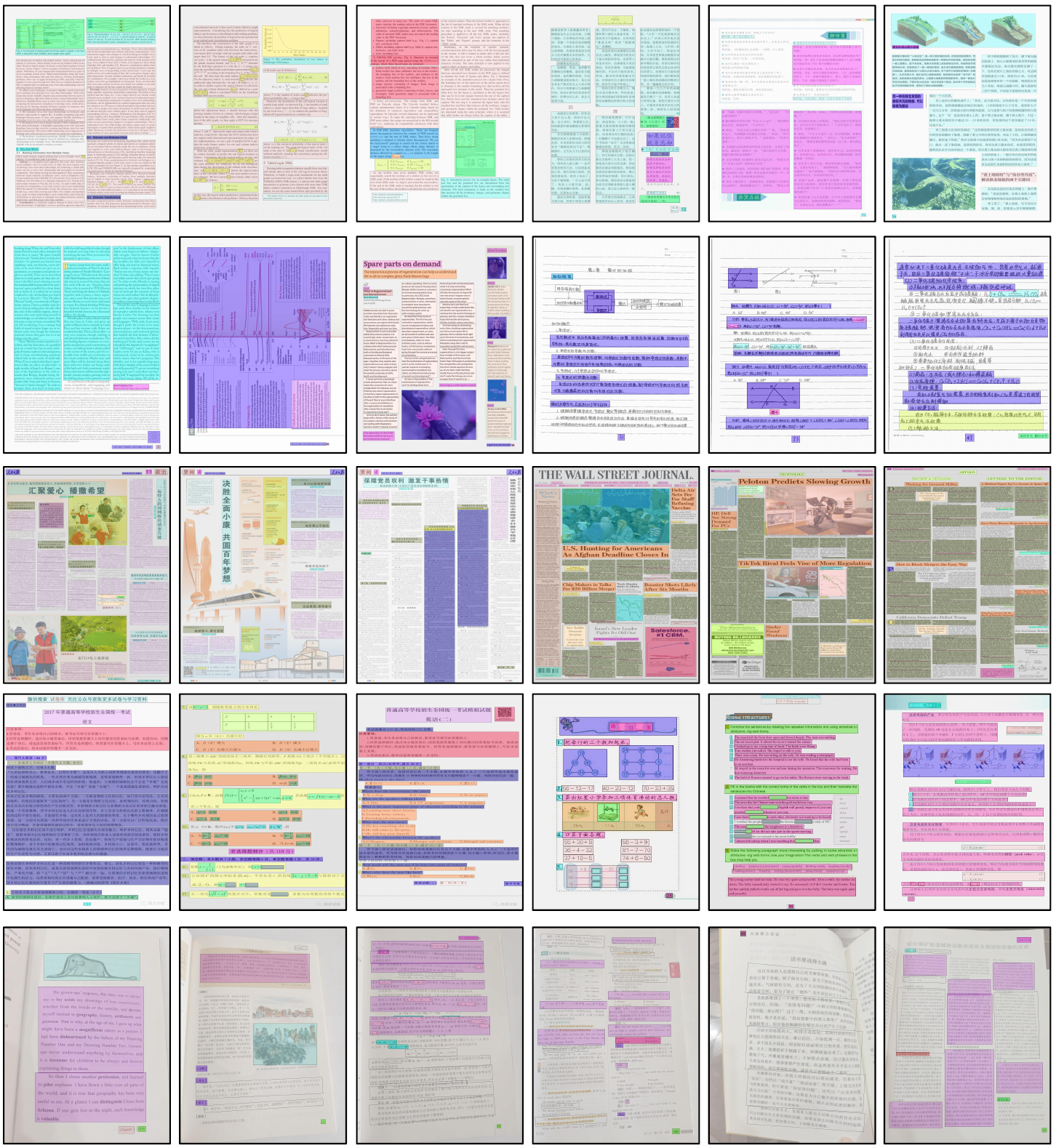}
    \caption{The Visualization results of TransDLANet trained on and inferred on nine subsets. Zoom in for better view.}
    \label{visualization results}
\end{figure*}

In the following, we will specifically analyze the performance of TransDLANet on different datasets. Tables~\ref{scientific article result}, ~\ref{magazine ch result}, ~\ref{magazine en result}, ~\ref{note result}, ~\ref{newspaper ch result}, ~\ref{newspaper en result}, ~\ref{test paper result}, ~\ref{textbook result}, and ~\ref{book result} show the APs of different categories in each of the nine subsets. Figure~\ref{visualization results} shows the visualization results of TransDLANet trained on and inferred on nine subsets.

As shown in Table~\ref{scientific article result}, we can see that TransDLANet obtains low precision in the formula, ordered list, and table note categories in the scientific article subset. The visualization results show that the formula category's low precision is attributed to many small formulas that are not detected, and the low precision of the ordered list is due to its easy identification as a paragraph category. The low precision of the page number category is due to the small target of this category, and the insufficient resolution during training and inference. We can consider increasing the resolution of the images during training and inference to improve this phenomenon.

As shown in Table~\ref{magazine ch result}, the advertisement, byline, ordered list, poem, and table categories are almost unrecognizable. This is due to the small training set of these categories in the magazine\_ch subset, which contains only 2, 12, 2, 15, and 5 training samples, respectively. Therefore, the model cannot learn the characteristics of these categories well. This phenomenon is alleviated in the total dataset of $M^{6}Doc$ mixed with nine subsets. From the visualization results, we can see that the unordered list is confused during inference. This is because the training samples of the unordered list are small, and most of them are annotated with large areas, which leads to overfitting when the model learns this category. We can also see that the poem category may be recognized as the unordered list category because them has a uniform hanging indent format. Additionally, since the layout logic of ``China National Geographic'' is very different from other Chinese magazines, especially in the headline and subhead category, incorrect detection of headlines appears more frequently in this magazine.

As shown in Table~\ref{magazine en result}, the TransDLANet model can process the magazine\_en dataset well, and the accuracy of each category is relatively balanced, except for the footnote category. We can see from the visualization results that the main reason for the low accuracy of the footnote category is that the large footnote category cannot be identified, and we can also see that the current model is not able to handle rotated images. We can consider more data enhancements to adapt to this scenario.

As shown in Table~\ref{note result}, since the note subset contains handwritten notes, it does not have apparent font size and color differences like published printed documents, and there is also great individual variability in personal notes. Hence, almost all models fail to obtain better results. In particular, the ``figure'' category's accuracy is much lower than other subsets. This is because the main difference between handwritten and printed documents is that figures in the former are drawn by hand, which led to no significantly different color features between figure, paragraph, and table instances. The visualization results show that the formulas of both mathematics and chemistry are almost unrecognizable, probably because the format of handwritten formulas is not as strict as that of published printed documents.

As shown in Tables~\ref{newspaper ch result} and ~\ref{newspaper en result}, the model trained on the newspaper\_ch subset cannot identify the byline and index categories, while the newspaper\_en subset cannot identify the ordered list and unordered list categories. This is mainly because the annotated samples of these categories are too limited. From the visualization results, we can see that newspapers of this type have very dense text, so there are more missed detections of paragraph instances than in other subsets. This can be improved by increasing the number of pre-defined queries or the number of epochs for training. Moreover, we can think about how to solve the problem of low recall with query-based methods.

As shown in Table~\ref{test paper result}, the TransDLANet model can accurately identify and locate different categories of layout elements in educational documents. This phenomenon can be attributed to the relatively low ambiguity and confusion among categories in the layout analysis of educational documents. However, TransDLANet still faces some limitations and challenges. For example, the low precision of categories with fewer annotation samples, such as seal lines, unordered lists, candidate information areas, footnotes, and other question numbers, with training set sizes of 3, 5, 8, 30, and 42, respectively. This is because the sample distribution of different categories in the dataset is imbalanced, so some categories have very few samples, which results in poor performance of the model in these categories. Therefore, future research should focus on those categories with fewer samples to improve the performance of the model.

As shown in Table~\ref{textbook result}, the second-level title can almost not be recognized, and the third-level question number and fourth-level title are incorrectly identified as ordered lists or paragraphs because they are similar to ordered lists and the training sample size is small. This suggests that the performance of the model is still inadequate for labels with a small number of training samples and that more samples and better data enhancement methods are needed to improve its recognition performance.

Because the book subset comes from 50 books and only six graphs are trained for each book, processing this subset well is very difficult. As shown in Tables~\ref{tab: nine subsets results}, and  ~\ref{book result}, all the models cannot obtain good results, and almost all the categories cannot obtain high AP. In addition, the reason for the AP of the ``drop cap'' category being nan is that there is only one drop cap annotation in the book subset, which cannot be equally assigned to training, evaluation, and test sets. However, to maintain consistent annotation with other datasets, we kept this category. The visualization results are shown in Figure~\ref{visualization results}.

\begin{table}[!htbp]
    \caption{The performance of TransDLANet on the scientific article test set.}
    \label{scientific article result}
    \resizebox{\linewidth}{!}{
    \begin{tabular}{ll|ll|ll}
    \hline
    \textbf{Category} & \textbf{AP} & \textbf{Category} & \textbf{AP} & \textbf{Category} & \textbf{AP} \\ \hline
    algorithm & 84.379 & author & 42.522 & caption & 75.633 \\
    code & 50.000 & drop cap & 60.000 & figure & 79.765 \\
    footer & 50.217 & footnote & 66.101 & formula & 26.691 \\
    fourth-level section title & 48.513 & header & 66.968 & institute & 57.636 \\
    marginal note & 78.634 & ordered list & 21.704 & page number & 38.226 \\
    paragraph & 87.656 & reference & 94.281 & section title & 58.769 \\
    sub section title & 62.701 & subsub section title & 31.692 & table & 90.409 \\
    table caption & 64.005 & table note & 22.038 & title & 68.761 \\
    unordered list & 65.143 &  &  &  & \\ \hline
    \end{tabular}}
\end{table}
\begin{table}[!htbp]
    \caption{The performance of TransDLANet on the magazine\_ch test set.}
    \label{magazine ch result}
    \resizebox{\linewidth}{!}{
    \begin{tabular}{ll|ll|ll}
    \hline
    \textbf{Category} & \textbf{AP} & \textbf{Category} & \textbf{AP} & \textbf{Category} & \textbf{AP} \\ \hline
    QR code & 76.832 & advertisement & 0.130 & author & 54.551 \\
    byline & 0.652 & caption & 50.815 & credit & 73.215 \\
    figure & 77.505 & footer & 73.403 & header & 79.318 \\
    headline & 70.126 & ordered list & 0.009 & page number & 64.164 \\
    paragraph & 91.514 & poem & 0.000 & section & 74.005 \\
    subhead & 59.667 & supplementary note & 78.409 & table & 0.125 \\
    translator & 53.897 & unordered list & 25.326 &  & \\ \hline
    \end{tabular}}
\end{table}

\begin{table}[!htbp]
    \caption{The performance of TransDLANet on the magazine\_en test set.}
    \label{magazine en result}
    \resizebox{\linewidth}{!}{
    \begin{tabular}{ll|ll|ll}
    \hline
    \textbf{Category} & \textbf{AP} & \textbf{Category} & \textbf{AP} & \textbf{Category} & \textbf{AP} \\ \hline
    QR code & 80.000 & advertisement & 72.343 & author & 48.984 \\
    breakout & 83.364 & byline & 45.858 & caption & 68.649 \\
    credit & 60.942 & dateline & 61.780 & drop cap & 66.936 \\
    figure & 79.911 & footer & 72.285 & footnote & 35.439 \\
    header & 79.068 & headline & 72.863 & lead & 82.744 \\
    ordered list & 72.376 & page number & 59.108 & paragraph & 92.683 \\
    poem & 50.495 & section & 69.735 & sidebar & 65.461 \\
    subhead & 62.719 & supplementary note & 60.192 & unordered list & 91.683 \\ \hline
    \end{tabular}}
\end{table}

\begin{table}[!htbp]
    \caption{The performance of TransDLANet on the note test set.}
    \label{note result}
    \resizebox{\linewidth}{!}{
    \begin{tabular}{ll|ll|ll}
    \hline
    \textbf{Category} & \textbf{AP} & \textbf{Category} & \textbf{AP} & \textbf{Category} & \textbf{AP} \\ \hline
    answer & 44.210 & caption & 21.512 & catalogue & 87.947 \\
    chapter title & 58.599 & figure & 23.948 & footer & 70.653 \\
    formula & 13.688 & option & 43.396 & ordered list & 25.549 \\
    page number & 55.228 & paragraph & 55.782 & part & 21.201 \\
    section & 14.880 & section title & 59.468 & sub section title & 48.215 \\
    supplementary note & 0.332 & table & 91.555 & unordered list & 58.368 \\ \hline
    \end{tabular}}
\end{table}

\begin{table}[!htbp]
    \caption{The performance of TransDLANet on the newspaper\_ch test set.}
    \label{newspaper ch result}
    \resizebox{\linewidth}{!}{
    \begin{tabular}{ll|ll|ll}
    \hline
    \textbf{Category} & \textbf{AP} & \textbf{Category} & \textbf{AP} & \textbf{Category} & \textbf{AP} \\ \hline
    QR code & 58.442 & advertisement & 63.137 & author & 46.547 \\
    byline & 0.000 & caption & 43.161 & credit & 34.098 \\
    dateline & 59.176 & editor's note & 50.666 & figure & 69.642 \\
    flag & 97.481 & folio & 69.364 & footer & 63.756 \\
    headline & 73.578 & index & 0.129 & jump line & 54.670 \\
    kicker & 61.401 & lead & 77.944 & mugshot & 74.719 \\
    ordered list & 26.165 & page number & 50.256 & paragraph & 85.700 \\
    section & 79.598 & sidebar & 94.422 & subhead & 62.211 \\
    supplementary note & 66.535 & teasers & 86.906 & unordered list & 53.598 \\ \hline
    \end{tabular}}
\end{table}

\begin{table}[!htbp]
    \caption{The performance of TransDLANet on the newspaper\_en test set.}
    \label{newspaper en result}
    \resizebox{\linewidth}{!}{
    \begin{tabular}{ll|ll|ll}
    \hline
    \textbf{Category} & \textbf{AP} & \textbf{Category} & \textbf{AP} & \textbf{Category} & \textbf{AP} \\ \hline
    advertisement & 73.364 & author & 49.603 & barcode & 86.667 \\
    bill & 93.366 & breakout & 82.382 & byline & 63.751 \\
    caption & 63.347 & correction & 19.507 & credit & 54.481 \\
    dateline & 41.799 & drop cap & 76.069 & figure & 81.021 \\
    flag & 94.422 & folio & 76.453 & headline & 83.541 \\
    index & 80.572 & inside & 86.799 & jump line & 65.844 \\
    kicker & 48.362 & lead & 74.842 & mugshot & 69.183 \\
    ordered list & 2.106 & page number & 68.168 & paragraph & 91.322 \\
    play & 100.000 & section & 79.960 & sidebar & 31.977 \\
    subhead & 58.967 & supplementary note & 54.872 & table & 34.636 \\
    table caption & 14.622 & teasers & 74.455 & unordered list & 0.000 \\
    weather forecast & 100.000 &  &  &  &  \\ \hline
    \end{tabular}}
\end{table}

\begin{table}[!htbp]
    \caption{The performance of TransDLANet on the test paper test set.}
    \label{test paper result}
    \resizebox{\linewidth}{!}{
    \begin{tabular}{ll|ll|ll}
    \hline
    \textbf{Category} & \textbf{AP} & \textbf{Category} & \textbf{AP} & \textbf{Category} & \textbf{AP} \\ \hline
    QR code & 91.122 & author & 59.273 & bracket & 69.809 \\
    byline & 71.388 & caption & 51.183 & endnote & 30.768 \\
    examinee information & 14.891 & figure & 74.224 & first-level question number & 63.711 \\
    first-level title & 76.011 & footer & 72.801 & formula & 54.910 \\
    header & 76.675 & headline & 59.850 & option & 85.754 \\
    ordered list & 75.898 & other question number & 19.441 & page number & 59.033 \\
    paragraph & 80.328 & part & 71.907 & poem & 82.214 \\
    sealing line & 24.158 & second-level question number & 65.567 & second-level title & 76.418 \\
    supplementary note & 63.752 & table & 91.735 & table caption & 39.857 \\
    third-level question number & 43.966 & title & 81.686 & underscore & 53.332 \\
    unordered list & 0.108 &  &  &  & \\ \hline
    \end{tabular}}
\end{table}

\begin{table}[!htbp]
    \caption{The performance of TransDLANet on the textbook test set.}
    \label{textbook result}
    \resizebox{\linewidth}{!}{
    \begin{tabular}{ll|ll|ll}
    \hline
    \textbf{Category} & \textbf{AP} & \textbf{Category} & \textbf{AP} & \textbf{Category} & \textbf{AP} \\ \hline
    answer & 17.774 & author & 30.492 & blank & 41.271 \\
    bracket & 64.726 & byline & 30.133 & caption & 65.620 \\
    catalogue & 42.970 & chapter title & 67.027 & credit & 22.500 \\
    dateline & 35.842 & figure & 69.836 & first-level question number & 50.457 \\
    first-level title & 73.183 & footer & 74.969 & footnote & 78.615 \\
    formula & 36.004 & fourth-level title & 18.654 & header & 72.671 \\
    headline & 71.452 & index & 66.639 & lead & 81.441 \\
    marginal note & 77.873 & matching & 22.620 & option & 50.744 \\
    ordered list & 22.818 & page number & 58.093 & paragraph & 73.979 \\
    part & 73.818 & poem & 29.581 & second-level question number & 56.581 \\
    second-level title & 0.345 & section & 74.510 & section title & 72.277 \\
    sub section title & 69.981 & subhead & 53.934 & supplementary note & 24.827 \\
    table & 76.178 & table caption & 54.939 & third-level question number & 0.000 \\
    third-level title & 46.164 & underscore & 54.956 & unordered list & 63.386 \\ \hline
    \end{tabular}}
\end{table}

\begin{table}[!htbp]
    \caption{The performance of TransDLANet on the book test set.}
    \label{book result}
    \resizebox{\linewidth}{!}{
    \begin{tabular}{ll|ll|ll}
    \hline
    \textbf{Category} & \textbf{AP} & \textbf{Category} & \textbf{AP} & \textbf{Category} & \textbf{AP} \\ \hline
    QR code & 0.000 & answer & 0.000 & author & 28.297 \\
    bracket & 19.010 & byline & 2.574 & caption & 40.736 \\
    catalogue & 0.000 & chapter title & 21.729 & code & 30.138 \\
    drop cap & nan & endnote & 10.575 & fifth-level title & 0.085 \\
    figure & 55.424 & first-level question number & 23.994 & first-level title & 46.635 \\
    footer & 47.277 & footnote & 0.000 & formula & 21.660 \\
    fourth-level title & 7.271 & header & 69.067 & headline & 25.235 \\
    index & 83.414 & institute & 47.772 & jump line & 0.000 \\
    marginal note & 52.954 & option & 17.763 & ordered list & 10.103 \\
    page number & 56.186 & paragraph & 60.591 & poem & 0.000 \\
    reference & 64.406 & second-level question number & 9.172 & second-level title & 0.671 \\
    section & 38.086 & section title & 40.258 & sub section title & 42.836 \\
    subsub section title & 24.022 & supplementary note & 0.000 & table & 58.094 \\
    table caption & 41.820 & third-level title & 28.857 & title & 75.264 \\
    underscore & 0.000 & unordered list & 14.470 &  & \\ \hline
    \end{tabular}}
\end{table}

\section{Discussion of Failure Cases of DocLayNet and PubLayNet datasets}
\begin{figure}[!htbp]
    \centering
        \includegraphics[width=\linewidth]{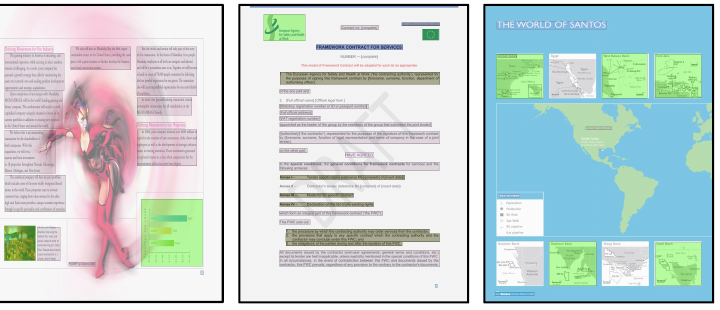}
    \caption{Failure Cases of DocLayNet datasets.}
    \label{Erroneous_DocLayNet}
\end{figure}

\begin{figure}[!htbp]
    \centering
        \includegraphics[width=\linewidth]{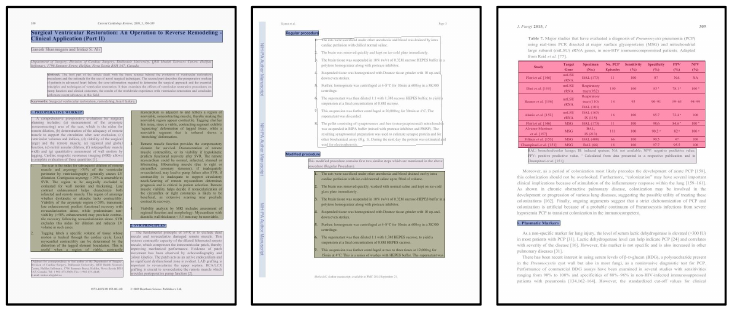}
    \caption{Failure Cases of PubLayNet datasets.}
    \label{Erroneous_PubLayNet}
\end{figure}

Figures~\ref{Erroneous_DocLayNet}, ~\ref{Erroneous_PubLayNet} show some of the failures of the TransDLANet model on the DocLayNet and PubLayNet datasets.

As shown in Figures~\ref{Erroneous_DocLayNet} and~\ref{Erroneous_PubLayNet}, the TransDLANet model trained on both the DocLayNet and PubLayNet datasets suffers from a missing detection problem during inference. The main reason for the low overall accuracy of our model is that we have fixed the number of queries in advance. Therefore, if there are multiple queries corresponding to a single instance, our model may fail to detect all the instances in the images. It is worth noting that the visualization results of PubLayNet inference also show some tilt detection errors, which may be caused by noise in the PubLayNet dataset.

\section{Annotation samples of $M^{6}Doc$}
\begin{figure*}[!ht]
    \centering
    \includegraphics[width=\textwidth]{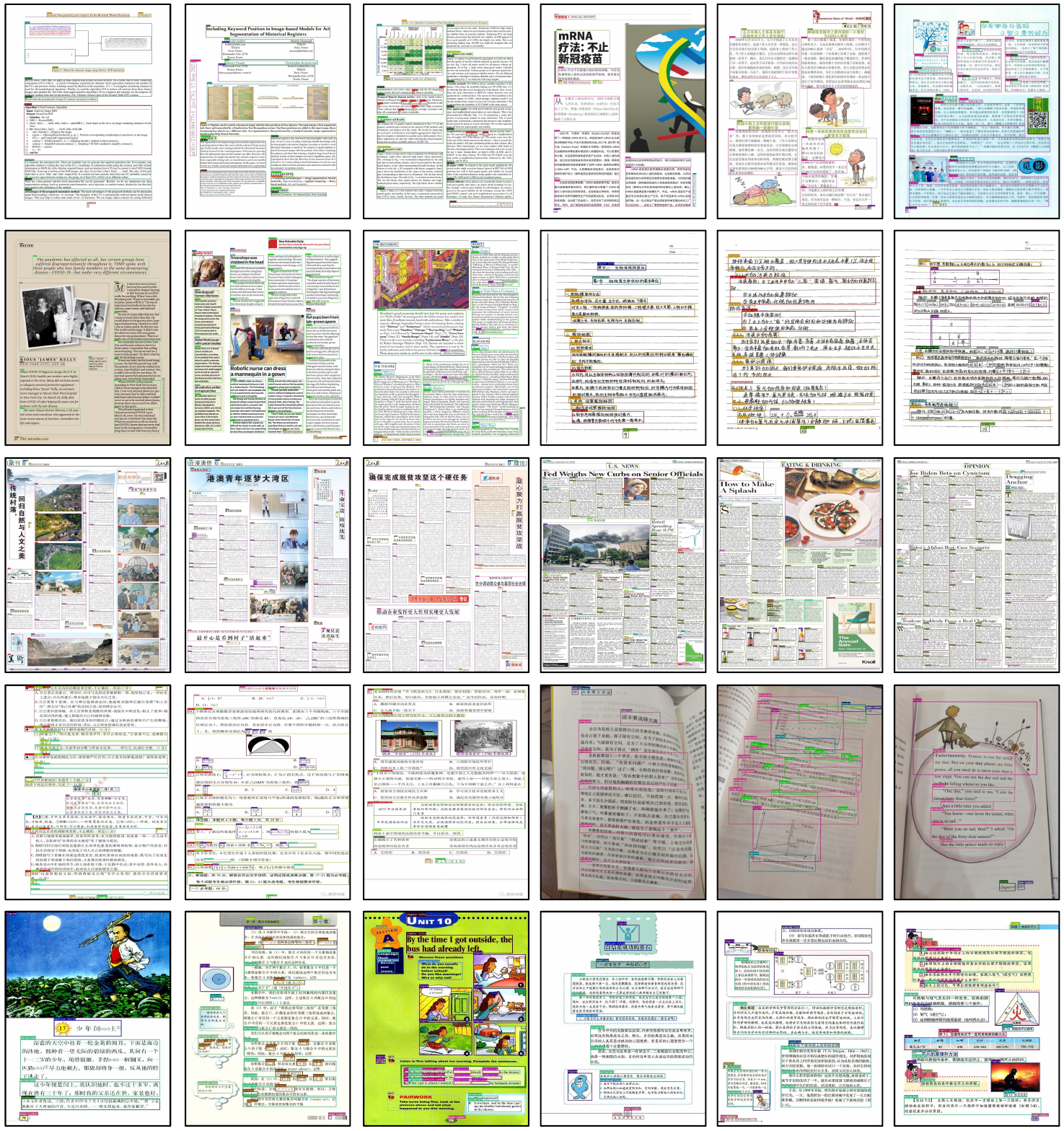}
    \caption{Example annotations of the $M^{6}Doc$. Zoom in for better view.}
    \label{Annotations}
\end{figure*}

Figure~\ref{Annotations} shows annotation samples of $M^{6}Doc$ dataset. There are a total of 74 annotation categories in our dataset. Among them, scientific article, textbook, book, test paper, magazine\_ch, magazine\_en, newspaper\_ch, newspaer\_en, and note subsets with 25, 42, 44, 31, 22, 24, 27, 34, and 18 categories, respectively. Detailed category statistics of the training, validation, and test sets for the nine subsets are shown in Tables~\ref{tab: scientific article subset}, ~\ref{tab: textbook subset}, ~\ref{tab: newspaper ch subset}, ~\ref{tab: newspaper en subset}, ~\ref{tab: book subset}, ~\ref{tab: magazine ch subset}, ~\ref{tab: test paper subset}, ~\ref{tab: magazine en subset}, and ~\ref{tab: note subset}. 

\begin{table}[!htbp]
\caption{Scientific article subset overview.}
\label{tab: scientific article subset}
\resizebox{\linewidth}{!}{
    \begin{tabular}{l|llllll}
    \hline
    \multicolumn{1}{l|}{\multirow{2}{*}{\textbf{Category}}} & \multicolumn{2}{c}{\textbf{Training}} & \multicolumn{2}{c}{\textbf{Validate}} & \multicolumn{2}{c}{\textbf{Testing}} \\
    \multicolumn{1}{c|}{} & \textbf{Number} & \textbf{\%} & \textbf{Number} & \textbf{\%} & \textbf{Number} & \textbf{\%} \\ \hline
    algorithm & 12 & 0.14 & 3 & 0.21 & 12 & 0.28 \\
    author & 51 & 0.62 & 8 & 0.56 & 29 & 0.67 \\
    caption & 371 & 4.48 & 64 & 4.51 & 174 & 4.03 \\
    code & 2 & 0.02 & 2 & 0.14 & 1 & 0.02 \\
    drop cap & 2 & 0.02 & 1 & 0.07 & 1 & 0.02 \\
    figure & 368 & 4.44 & 63 & 4.44 & 184 & 4.26 \\
    footer & 32 & 0.39 & 7 & 0.49 & 17 & 0.39 \\
    footnote & 128 & 1.54 & 18 & 1.27 & 65 & 1.51 \\
    formula & 2081 & 25.11 & 363 & 25.60 & 1106 & 25.62 \\
    fourth-level section title & 15 & 0.18 & 3 & 0.21 & 19 & 0.44 \\
    header & 219 & 2.64 & 31 & 2.19 & 131 & 3.03 \\
    institute & 49 & 0.59 & 8 & 0.56 & 24 & 0.56 \\
    marginal note & 40 & 0.48 & 5 & 0.35 & 20 & 0.46 \\
    ordered list & 54 & 0.65 & 12 & 0.85 & 38 & 0.88 \\
    page number & 369 & 4.45 & 59 & 4.16 & 175 & 4.05 \\
    paragraph & 2916 & 35.18 & 522 & 36.81 & 1551 & 35.93 \\
    reference & 134 & 1.62 & 20 & 1.41 & 57 & 1.32 \\
    section title & 432 & 5.21 & 76 & 5.36 & 221 & 5.12 \\
    sub section title & 412 & 4.97 & 71 & 5.01 & 191 & 4.42 \\
    subsub section title & 80 & 0.97 & 17 & 1.20 & 48 & 1.11 \\
    table & 181 & 2.18 & 22 & 1.55 & 88 & 2.04 \\
    table caption & 177 & 2.14 & 19 & 1.34 & 80 & 1.85 \\
    table note & 8 & 0.10 & 2 & 0.14 & 5 & 0.12 \\
    title & 46 & 0.56 & 6 & 0.42 & 25 & 0.58 \\
    unordered list & 109 & 1.32 & 16 & 1.13 & 55 & 1.27 \\
    \textbf{Total} & \textbf{8288} & \textbf{100} & \textbf{1418} & \textbf{100} & \textbf{4317} & \textbf{100} \\ \hline
    \end{tabular}}
\end{table}
\begin{table}[!htbp]
\caption{Textbook subset overview.}
\label{tab: textbook subset}
\resizebox{\linewidth}{!}{
    \begin{tabular}{l|llllll}
    \hline
    \multicolumn{1}{l|}{\multirow{2}{*}{\textbf{Category}}} & \multicolumn{2}{c}{\textbf{Training}} & \multicolumn{2}{c}{\textbf{Validate}} & \multicolumn{2}{c}{\textbf{Testing}} \\
    \multicolumn{1}{c|}{} & \textbf{Number} & \textbf{\%} & \textbf{Number} & \textbf{\%} & \textbf{Number} & \textbf{\%} \\ \hline
    answer & 27 & 0.12 & 8 & 0.22 & 20 & 0.18 \\
    author & 25 & 0.11 & 7 & 0.19 & 13 & 0.12 \\
    blank & 189 & 0.81 & 58 & 1.59 & 90 & 0.80 \\
    bracket & 237 & 1.01 & 34 & 0.93 & 74 & 0.66 \\
    byline & 34 & 0.15 & 8 & 0.22 & 32 & 0.29 \\
    caption & 867 & 3.71 & 157 & 4.31 & 423 & 3.77 \\
    catalogue & 7 & 0.03 & 3 & 0.08 & 4 & 0.04 \\
    chapter title & 164 & 0.70 & 23 & 0.63 & 82 & 0.73 \\
    credit & 1 & 0.00 & 1 & 0.03 & 1 & 0.01 \\
    dateline & 11 & 0.05 & 3 & 0.08 & 3 & 0.03 \\
    figure & 2239 & 9.57 & 375 & 10.29 & 1106 & 9.87 \\
    first-level question number & 1634 & 6.98 & 245 & 6.72 & 695 & 6.20 \\
    first-level title & 158 & 0.68 & 15 & 0.41 & 74 & 0.66 \\
    footer & 480 & 2.05 & 85 & 2.33 & 241 & 2.15 \\
    footnote & 158 & 0.68 & 26 & 0.71 & 69 & 0.62 \\
    formula & 2164 & 9.25 & 304 & 8.34 & 1031 & 9.20 \\
    fourth-level title & 7 & 0.03 & 8 & 0.22 & 15 & 0.13 \\
    header & 434 & 1.85 & 64 & 1.76 & 224 & 2.00 \\
    headline & 494 & 2.11 & 91 & 2.50 & 240 & 2.14 \\
    index & 97 & 0.41 & 15 & 0.41 & 47 & 0.42 \\
    lead & 46 & 0.20 & 12 & 0.33 & 22 & 0.20 \\
    marginal note & 165 & 0.71 & 29 & 0.80 & 75 & 0.67 \\
    matching & 7 & 0.03 & 1 & 0.03 & 8 & 0.07 \\
    option & 110 & 0.47 & 12 & 0.33 & 33 & 0.29 \\
    ordered list & 159 & 0.68 & 22 & 0.60 & 71 & 0.63 \\
    page number & 1183 & 5.06 & 193 & 5.29 & 590 & 5.26 \\
    paragraph & 8223 & 35.14 & 1245 & 34.15 & 3999 & 35.68 \\
    part & 412 & 1.76 & 70 & 1.92 & 219 & 1.95 \\
    poem & 39 & 0.17 & 9 & 0.25 & 19 & 0.17 \\
    second-level question number & 676 & 2.89 & 48 & 1.32 & 295 & 2.63 \\
    second-level title & 2 & 0.01 & 1 & 0.03 & 1 & 0.01 \\
    section & 921 & 3.94 & 153 & 4.20 & 470 & 4.19 \\
    section title & 223 & 0.95 & 51 & 1.40 & 88 & 0.79 \\
    sub section title & 31 & 0.13 & 6 & 0.16 & 14 & 0.12 \\
    subhead & 93 & 0.40 & 22 & 0.60 & 43 & 0.38 \\
    supplementary note & 21 & 0.09 & 2 & 0.05 & 16 & 0.14 \\
    table & 273 & 1.17 & 51 & 1.40 & 136 & 1.21 \\
    table caption & 44 & 0.19 & 5 & 0.14 & 16 & 0.14 \\
    third-level question number & 11 & 0.05 & 1 & 0.03 & 7 & 0.06 \\
    third-level title & 50 & 0.21 & 18 & 0.49 & 35 & 0.31 \\
    underscore & 1066 & 4.56 & 125 & 3.43 & 438 & 3.91 \\
    unordered list & 217 & 0.93 & 40 & 1.10 & 130 & 1.16 \\
    \textbf{Total} & \textbf{23399} & \textbf{100} & \textbf{3646} & \textbf{100} & \textbf{11209} & \textbf{100} \\ \hline
    \end{tabular}}
\end{table}
\begin{table}[!htbp]
\caption{Newspaper\_ch subset overview.}
\label{tab: newspaper ch subset}
\resizebox{\linewidth}{!}{
    \begin{tabular}{l|llllll}
    \hline
    \multicolumn{1}{l|}{\multirow{2}{*}{\textbf{Category}}} & \multicolumn{2}{c}{\textbf{Training}} & \multicolumn{2}{c}{\textbf{Validate}} & \multicolumn{2}{c}{\textbf{Testing}} \\
    \multicolumn{1}{c|}{} & \textbf{Number} & \textbf{\%} & \textbf{Number} & \textbf{\%} & \textbf{Number} & \textbf{\%} \\ \hline
    QR code & 12 & 0.05 & 6 & 0.16 & 11 & 0.09 \\
    advertisement & 28 & 0.12 & 10 & 0.27 & 21 & 0.18 \\
    author & 1622 & 7.03 & 260 & 7.03 & 808 & 6.91 \\
    byline & 2 & 0.01 & 1 & 0.03 & 11 & 0.09 \\
    caption & 375 & 1.63 & 70 & 1.89 & 199 & 1.70 \\
    credit & 390 & 1.69 & 73 & 1.97 & 201 & 1.72 \\
    dateline & 562 & 2.44 & 87 & 2.35 & 277 & 2.37 \\
    editor's note & 39 & 0.17 & 4 & 0.11 & 9 & 0.08 \\
    figure & 797 & 3.46 & 124 & 3.35 & 389 & 3.33 \\
    flag & 20 & 0.09 & 4 & 0.11 & 9 & 0.08 \\
    folio & 598 & 2.59 & 100 & 2.70 & 301 & 2.57 \\
    footer & 224 & 0.97 & 18 & 0.49 & 90 & 0.77 \\
    headline & 1327 & 5.75 & 212 & 5.73 & 643 & 5.50 \\
    index & 4 & 0.02 & 1 & 0.03 & 3 & 0.03 \\
    jump line & 128 & 0.56 & 19 & 0.51 & 54 & 0.46 \\
    kicker & 359 & 1.56 & 68 & 1.84 & 169 & 1.45 \\
    lead & 193 & 0.84 & 34 & 0.92 & 88 & 0.75 \\
    mugshot & 13 & 0.06 & 4 & 0.11 & 6 & 0.05 \\
    ordered list & 83 & 0.36 & 17 & 0.46 & 40 & 0.34 \\
    page number & 280 & 1.21 & 46 & 1.24 & 140 & 1.20 \\
    paragraph & 14520 & 62.96 & 2281 & 61.63 & 7430 & 63.56 \\
    section & 284 & 1.23 & 46 & 1.24 & 142 & 1.21 \\
    sidebar & 3 & 0.01 & 3 & 0.08 & 3 & 0.03 \\
    subhead & 1034 & 4.48 & 188 & 5.08 & 573 & 4.90 \\
    supplementary note & 146 & 0.63 & 21 & 0.57 & 65 & 0.56 \\
    teasers & 12 & 0.05 & 1 & 0.03 & 4 & 0.03 \\
    unordered list & 8 & 0.03 & 3 & 0.08 & 4 & 0.03 \\
    \textbf{Total} & \textbf{23063} & \textbf{100} & \textbf{3701} & \textbf{100} & \textbf{11690} & \textbf{100} \\ \hline
    \end{tabular}}
    \end{table}
\begin{table}[!htbp]
\caption{Newspaper\_en subset overview.}
\label{tab: newspaper en subset}
\resizebox{\linewidth}{!}{
    \begin{tabular}{l|llllll}
    \hline
    \multicolumn{1}{l|}{\multirow{2}{*}{\textbf{Category}}} & \multicolumn{2}{c}{\textbf{Training}} & \multicolumn{2}{c}{\textbf{Validate}} & \multicolumn{2}{c}{\textbf{Testing}} \\
    \multicolumn{1}{c|}{} & \textbf{Number} & \textbf{\%} & \textbf{Number} & \textbf{\%} & \textbf{Number} & \textbf{\%} \\ \hline
    advertisement & 154 & 0.73 & 25 & 0.75 & 92 & 0.88 \\
    author & 133 & 0.63 & 28 & 0.84 & 66 & 0.63 \\
    barcode & 10 & 0.05 & 1 & 0.03 & 3 & 0.03 \\
    bill & 3 & 0.01 & 2 & 0.06 & 3 & 0.03 \\
    breakout & 298 & 1.42 & 49 & 1.47 & 139 & 1.34 \\
    byline & 839 & 3.99 & 109 & 3.26 & 369 & 3.55 \\
    caption & 634 & 3.02 & 97 & 2.90 & 340 & 3.27 \\
    correction & 9 & 0.04 & 1 & 0.03 & 6 & 0.06 \\
    credit & 567 & 2.70 & 97 & 2.90 & 278 & 2.67 \\
    dateline & 151 & 0.72 & 23 & 0.69 & 78 & 0.75 \\
    drop cap & 188 & 0.89 & 30 & 0.90 & 118 & 1.13 \\
    figure & 799 & 3.80 & 127 & 3.80 & 416 & 4.00 \\
    flag & 10 & 0.05 & 1 & 0.03 & 3 & 0.03 \\
    folio & 844 & 4.02 & 113 & 3.38 & 384 & 3.69 \\
    headline & 986 & 4.69 & 147 & 4.40 & 454 & 4.36 \\
    index & 83 & 0.39 & 13 & 0.39 & 30 & 0.29 \\
    inside & 16 & 0.08 & 1 & 0.03 & 5 & 0.05 \\
    jump line & 252 & 1.20 & 43 & 1.29 & 125 & 1.20 \\
    kicker & 157 & 0.75 & 23 & 0.69 & 88 & 0.85 \\
    lead & 201 & 0.96 & 29 & 0.87 & 71 & 0.68 \\
    mugshot & 60 & 0.29 & 7 & 0.21 & 40 & 0.38 \\
    ordered list & 6 & 0.03 & 4 & 0.12 & 4 & 0.04 \\
    page number & 290 & 1.38 & 49 & 1.47 & 149 & 1.43 \\
    paragraph & 13435 & 63.92 & 2142 & 64.13 & 6680 & 64.21 \\
    play & 10 & 0.05 & 3 & 0.09 & 2 & 0.02 \\
    section & 343 & 1.63 & 63 & 1.89 & 168 & 1.61 \\
    sidebar & 48 & 0.23 & 6 & 0.18 & 21 & 0.20 \\
    subhead & 215 & 1.02 & 30 & 0.90 & 96 & 0.92 \\
    supplementary note & 172 & 0.82 & 31 & 0.93 & 99 & 0.95 \\
    table & 55 & 0.26 & 25 & 0.75 & 43 & 0.41 \\
    table caption & 13 & 0.06 & 11 & 0.33 & 25 & 0.24 \\
    teasers & 20 & 0.10 & 6 & 0.18 & 3 & 0.03 \\
    unordered list & 6 & 0.03 & 1 & 0.03 & 3 & 0.03 \\
    weather forecast & 10 & 0.05 & 3 & 0.09 & 3 & 0.03 \\
    \textbf{Total} & \textbf{21017} & \textbf{100} & \textbf{3340} & \textbf{100} & \textbf{10404} & \textbf{100} \\ \hline
    \end{tabular}}
    \end{table}
\begin{table}[!htbp]
\caption{Book subset overview.}
\label{tab: book subset}
\resizebox{\linewidth}{!}{
    \begin{tabular}{l|llllll}
    \hline
    \multicolumn{1}{l|}{\multirow{2}{*}{\textbf{Category}}} & \multicolumn{2}{c}{\textbf{Training}} & \multicolumn{2}{c}{\textbf{Validate}} & \multicolumn{2}{c}{\textbf{Testing}} \\
    \multicolumn{1}{c|}{} & \textbf{Number} & \textbf{\%} & \textbf{Number} & \textbf{\%} & \textbf{Number} & \textbf{\%} \\ \hline
    QR code & 3 & 0.05 & 1 & 0.10 & 3 & 0.11 \\
    answer & 4 & 0.07 & 1 & 0.10 & 1 & 0.04 \\
    author & 15 & 0.27 & 1 & 0.10 & 4 & 0.15 \\
    bracket & 11 & 0.20 & 1 & 0.10 & 4 & 0.15 \\
    byline & 5 & 0.09 & 1 & 0.10 & 3 & 0.11 \\
    caption & 77 & 1.37 & 12 & 1.25 & 33 & 1.24 \\
    catalogue & 2 & 0.04 & 1 & 0.10 & 1 & 0.04 \\
    chapter title & 40 & 0.71 & 5 & 0.52 & 17 & 0.64 \\
    code & 60 & 1.07 & 5 & 0.52 & 30 & 1.13 \\
    drop cap & 1 & 0.02 & 0 & 0.00 & 0 & 0.00 \\
    endnote & 5 & 0.09 & 2 & 0.21 & 6 & 0.23 \\
    fifth-level title & 13 & 0.23 & 2 & 0.21 & 20 & 0.75 \\
    figure & 113 & 2.01 & 18 & 1.88 & 50 & 1.88 \\
    first-level question number & 156 & 2.78 & 39 & 4.06 & 51 & 1.92 \\
    first-level title & 24 & 0.43 & 3 & 0.31 & 23 & 0.86 \\
    footer & 50 & 0.89 & 8 & 0.83 & 27 & 1.01 \\
    footnote & 4 & 0.07 & 1 & 0.10 & 1 & 0.04 \\
    formula & 1330 & 23.67 & 250 & 26.04 & 527 & 19.80 \\
    fourth-level title & 63 & 1.12 & 5 & 0.52 & 51 & 1.92 \\
    header & 206 & 3.67 & 36 & 3.75 & 102 & 3.83 \\
    headline & 53 & 0.94 & 3 & 0.31 & 30 & 1.13 \\
    index & 30 & 0.53 & 7 & 0.73 & 20 & 0.75 \\
    institute & 11 & 0.20 & 1 & 0.10 & 4 & 0.15 \\
    jump line & 1 & 0.02 & 1 & 0.10 & 1 & 0.04 \\
    marginal note & 33 & 0.59 & 3 & 0.31 & 6 & 0.23 \\
    option & 15 & 0.27 & 1 & 0.10 & 6 & 0.23 \\
    ordered list & 141 & 2.51 & 19 & 1.98 & 76 & 2.86 \\
    page number & 273 & 4.86 & 45 & 4.69 & 140 & 5.26 \\
    paragraph & 2156 & 38.38 & 353 & 36.77 & 1050 & 39.46 \\
    poem & 6 & 0.11 & 1 & 0.10 & 1 & 0.04 \\
    reference & 15 & 0.27 & 3 & 0.31 & 5 & 0.19 \\
    second-level question number & 103 & 1.83 & 16 & 1.67 & 30 & 1.13 \\
    second-level title & 4 & 0.07 & 2 & 0.21 & 2 & 0.08 \\
    section & 148 & 2.63 & 23 & 2.40 & 78 & 2.93 \\
    section title & 99 & 1.76 & 13 & 1.35 & 59 & 2.22 \\
    sub section title & 79 & 1.41 & 21 & 2.19 & 35 & 1.32 \\
    subsub section title & 21 & 0.37 & 4 & 0.42 & 23 & 0.86 \\
    supplementary note & 3 & 0.05 & 0 & 0.00 & 1 & 0.04 \\
    table & 81 & 1.44 & 14 & 1.46 & 41 & 1.54 \\
    table caption & 26 & 0.46 & 3 & 0.31 & 11 & 0.41 \\
    third-level title & 96 & 1.71 & 26 & 2.71 & 59 & 2.22 \\
    title & 11 & 0.20 & 1 & 0.10 & 4 & 0.15 \\
    underscore & 15 & 0.27 & 3 & 0.31 & 10 & 0.38 \\
    unordered list & 16 & 0.28 & 5 & 0.52 & 15 & 0.56 \\
    \textbf{Total} & \textbf{5618} & \textbf{100} & \textbf{960} & \textbf{100} & \textbf{2661} & \textbf{100} \\ \hline
    \end{tabular}}
\end{table}
\begin{table}[!htbp]
\caption{Magazine\_ch subset overview.}
\label{tab: magazine ch subset}
\resizebox{\linewidth}{!}{
    \begin{tabular}{l|llllll}
    \hline
    \multicolumn{1}{l|}{\multirow{2}{*}{\textbf{Category}}} & \multicolumn{2}{c}{\textbf{Training}} & \multicolumn{2}{c}{\textbf{Validate}} & \multicolumn{2}{c}{\textbf{Testing}} \\
    \multicolumn{1}{c|}{} & \textbf{Number} & \textbf{\%} & \textbf{Number} & \textbf{\%} & \textbf{Number} & \textbf{\%} \\ \hline
    QR code & 33 & 0.29 & 6 & 0.32 & 5 & 0.09 \\
    advertisement & 2 & 0.02 & 1 & 0.05 & 1 & 0.02 \\
    author & 421 & 3.71 & 73 & 3.89 & 182 & 3.28 \\
    byline & 12 & 0.11 & 2 & 0.11 & 37 & 0.67 \\
    caption & 397 & 3.50 & 53 & 2.82 & 206 & 3.72 \\
    credit & 271 & 2.39 & 42 & 2.24 & 120 & 2.17 \\
    figure & 1020 & 9.00 & 168 & 8.95 & 488 & 8.81 \\
    footer & 311 & 2.74 & 52 & 2.77 & 162 & 2.92 \\
    header & 369 & 3.25 & 56 & 2.98 & 181 & 3.27 \\
    headline & 491 & 4.33 & 82 & 4.37 & 214 & 3.86 \\
    ordered list & 2 & 0.02 & 1 & 0.05 & 5 & 0.09 \\
    page number & 569 & 5.02 & 95 & 5.06 & 285 & 5.14 \\
    paragraph & 6542 & 57.69 & 1076 & 57.33 & 3234 & 58.36 \\
    poem & 15 & 0.13 & 3 & 0.16 & 2 & 0.04 \\
    section & 218 & 1.92 & 33 & 1.76 & 107 & 1.93 \\
    subhead & 499 & 4.40 & 109 & 5.81 & 231 & 4.17 \\
    supplementary note & 60 & 0.53 & 9 & 0.48 & 28 & 0.51 \\
    table & 5 & 0.04 & 1 & 0.05 & 1 & 0.02 \\
    translator & 73 & 0.64 & 11 & 0.59 & 38 & 0.69 \\
    unordered list & 29 & 0.26 & 4 & 0.21 & 14 & 0.25 \\
    \textbf{Total} & \textbf{11339} & \textbf{100} & \textbf{1877} & \textbf{100} & \textbf{5541} & \textbf{100} \\ \hline
    \end{tabular}}
\end{table}
\begin{table}[!htbp]
\caption{Test paper subset overview.}
\label{tab: test paper subset}
\resizebox{\linewidth}{!}{
    \begin{tabular}{l|llllll}
    \hline
    \multicolumn{1}{l|}{\multirow{2}{*}{\textbf{Category}}} & \multicolumn{2}{c}{\textbf{Training}} & \multicolumn{2}{c}{\textbf{Validate}} & \multicolumn{2}{c}{\textbf{Testing}} \\
    \multicolumn{1}{c|}{} & \textbf{Number} & \textbf{\%} & \textbf{Number} & \textbf{\%} & \textbf{Number} & \textbf{\%} \\ \hline
    QR code & 4 & 0.01 & 1 & 0.02 & 3 & 0.02 \\
    author & 54 & 0.16 & 4 & 0.07 & 18 & 0.11 \\
    bracket & 615 & 1.79 & 129 & 2.33 & 195 & 1.16 \\
    byline & 114 & 0.33 & 18 & 0.32 & 75 & 0.45 \\
    caption & 445 & 1.29 & 97 & 1.75 & 223 & 1.32 \\
    endnote & 30 & 0.09 & 2 & 0.04 & 13 & 0.08 \\
    examinee information & 8 & 0.02 & 2 & 0.04 & 6 & 0.04 \\
    figure & 1373 & 3.99 & 220 & 3.97 & 681 & 4.05 \\
    first-level question number & 3879 & 11.28 & 646 & 11.64 & 1994 & 11.84 \\
    first-level title & 404 & 1.17 & 63 & 1.14 & 195 & 1.16 \\
    footer & 312 & 0.91 & 46 & 0.83 & 168 & 1.00 \\
    formula & 7064 & 20.54 & 1040 & 18.75 & 3319 & 19.71 \\
    header & 386 & 1.12 & 64 & 1.15 & 192 & 1.14 \\
    headline & 187 & 0.54 & 18 & 0.32 & 71 & 0.42 \\
    option & 3046 & 8.86 & 496 & 8.94 & 1530 & 9.09 \\
    ordered list & 314 & 0.91 & 57 & 1.03 & 152 & 0.90 \\
    other question number & 42 & 0.12 & 3 & 0.05 & 31 & 0.18 \\
    page number & 979 & 2.85 & 174 & 3.14 & 490 & 2.91 \\
    paragraph & 9268 & 26.94 & 1480 & 26.68 & 4602 & 27.34 \\
    part & 109 & 0.32 & 17 & 0.31 & 62 & 0.37 \\
    poem & 28 & 0.08 & 2 & 0.04 & 9 & 0.05 \\
    sealing line & 3 & 0.01 & 2 & 0.04 & 5 & 0.03 \\
    second-level question number & 1994 & 5.80 & 313 & 5.64 & 1005 & 5.97 \\
    second-level title & 267 & 0.78 & 45 & 0.81 & 137 & 0.81 \\
    supplementary note & 304 & 0.88 & 53 & 0.96 & 145 & 0.86 \\
    table & 157 & 0.46 & 27 & 0.49 & 65 & 0.39 \\
    table caption & 27 & 0.08 & 3 & 0.05 & 11 & 0.07 \\
    third-level question number & 229 & 0.67 & 35 & 0.63 & 95 & 0.56 \\
    title & 144 & 0.42 & 28 & 0.50 & 71 & 0.42 \\
    underscore & 2606 & 7.58 & 462 & 8.33 & 1269 & 7.54 \\
    unordered list & 5 & 0.01 & 1 & 0.02 & 3 & 0.02 \\
    \textbf{Total} & \textbf{34397} & \textbf{100} & \textbf{5548} & \textbf{100} & \textbf{16835} & \textbf{100} \\ \hline
    \end{tabular}}
\end{table}
\begin{table}[!htbp]
\caption{Magazine\_en subset overview.}
\label{tab: magazine en subset}
\resizebox{\linewidth}{!}{
    \begin{tabular}{l|llllll}
    \hline
    \multicolumn{1}{l|}{\multirow{2}{*}{\textbf{Category}}} & \multicolumn{2}{c}{\textbf{Training}} & \multicolumn{2}{c}{\textbf{Validate}} & \multicolumn{2}{c}{\textbf{Testing}} \\
    \multicolumn{1}{c|}{} & \textbf{Number} & \textbf{\%} & \textbf{Number} & \textbf{\%} & \textbf{Number} & \textbf{\%} \\ \hline
    QR code & 7 & 0.06 & 1 & 0.05 & 1 & 0.02 \\
    advertisement & 73 & 0.61 & 9 & 0.44 & 31 & 0.51 \\
    author & 103 & 0.87 & 22 & 1.07 & 68 & 1.12 \\
    breakout & 113 & 0.95 & 23 & 1.12 & 49 & 0.81 \\
    byline & 270 & 2.27 & 46 & 2.24 & 133 & 2.20 \\
    caption & 321 & 2.70 & 51 & 2.48 & 151 & 2.50 \\
    credit & 294 & 2.47 & 42 & 2.05 & 128 & 2.12 \\
    dateline & 177 & 1.49 & 27 & 1.32 & 124 & 2.05 \\
    drop cap & 223 & 1.88 & 40 & 1.95 & 115 & 1.90 \\
    figure & 843 & 7.10 & 137 & 6.67 & 410 & 6.78 \\
    footer & 474 & 3.99 & 79 & 3.85 & 235 & 3.89 \\
    footnote & 5 & 0.04 & 4 & 0.19 & 4 & 0.07 \\
    header & 263 & 2.21 & 46 & 2.24 & 139 & 2.30 \\
    headline & 577 & 4.86 & 90 & 4.38 & 329 & 5.44 \\
    lead & 224 & 1.89 & 34 & 1.66 & 104 & 1.72 \\
    ordered list & 25 & 0.21 & 1 & 0.05 & 7 & 0.12 \\
    page number & 535 & 4.50 & 90 & 4.38 & 265 & 4.38 \\
    paragraph & 6338 & 53.35 & 1134 & 55.24 & 3242 & 53.61 \\
    poem & 10 & 0.08 & 3 & 0.15 & 2 & 0.03 \\
    section & 565 & 4.76 & 88 & 4.29 & 254 & 4.20 \\
    sidebar & 3 & 0.03 & 1 & 0.05 & 3 & 0.05 \\
    subhead & 157 & 1.32 & 45 & 2.19 & 126 & 2.08 \\
    supplementary note & 272 & 2.29 & 39 & 1.90 & 124 & 2.05 \\
    unordered list & 9 & 0.08 & 1 & 0.05 & 3 & 0.05 \\
    \textbf{Total} & \multicolumn{1}{l}{\textbf{11881}} & \textbf{100} & \textbf{2053} & \textbf{100} & \textbf{6047} & \textbf{100} \\ \hline
    \end{tabular}}
\end{table}

\begin{table}[!htbp]
\caption{Note subset overview.}
\label{tab: note subset}
\resizebox{\linewidth}{!}{
    \begin{tabular}{l|llllll}
    \hline
    \multicolumn{1}{l|}{\multirow{2}{*}{\textbf{Category}}} & \multicolumn{2}{c}{\textbf{Training}} & \multicolumn{2}{c}{\textbf{Validate}} & \multicolumn{2}{c}{\textbf{Testing}} \\
    \multicolumn{1}{c|}{} & \textbf{Number} & \textbf{\%} & \textbf{Number} & \textbf{\%} & \textbf{Number} & \textbf{\%} \\ \hline
    answer & 134 & 3.32 & 21 & 3.15 & 56 & 2.59 \\
    caption & 21 & 0.52 & 4 & 0.60 & 17 & 0.79 \\
    catalogue & 30 & 0.74 & 6 & 0.90 & 14 & 0.65 \\
    chapter title & 41 & 1.02 & 5 & 0.75 & 25 & 1.16 \\
    figure & 62 & 1.54 & 10 & 1.50 & 38 & 1.76 \\
    footer & 101 & 2.50 & 15 & 2.25 & 47 & 2.17 \\
    formula & 451 & 11.17 & 101 & 15.14 & 208 & 9.62 \\
    option & 27 & 0.67 & 6 & 0.90 & 8 & 0.37 \\
    ordered list & 228 & 5.65 & 39 & 5.85 & 117 & 5.41 \\
    page number & 304 & 7.53 & 52 & 7.80 & 149 & 6.89 \\
    paragraph & 2244 & 55.57 & 342 & 51.27 & 1281 & 59.25 \\
    part & 3 & 0.07 & 2 & 0.30 & 2 & 0.09 \\
    section & 29 & 0.72 & 2 & 0.30 & 9 & 0.42 \\
    section title & 143 & 3.54 & 31 & 4.65 & 74 & 3.42 \\
    sub section title & 45 & 1.11 & 9 & 1.35 & 29 & 1.34 \\
    supplementary note & 8 & 0.20 & 3 & 0.45 & 9 & 0.42 \\
    table & 69 & 1.71 & 6 & 0.90 & 35 & 1.62 \\
    unordered list & 98 & 2.43 & 13 & 1.95 & 44 & 2.04 \\
    \textbf{Total} & \textbf{4038} & \textbf{100} & \textbf{667} & \textbf{100} & \textbf{2162} & \textbf{100} \\ \hline
    \end{tabular}}
\end{table}

\end{document}